\documentclass[letterpaper, 10 pt, conference]{ieeeconf}  

\IEEEoverridecommandlockouts                              

\usepackage{graphics} 
\usepackage{epsfig} 
\usepackage{times} 
\usepackage{amsmath} 
\usepackage{amssymb}  
\usepackage{bm}
\usepackage{color}
\usepackage{cite}
\usepackage[linesnumbered,ruled]{algorithm2e}
\usepackage{ulem} 
\usepackage{hyperref}
\usepackage{float}
\usepackage{booktabs}
\usepackage{multirow}
\usepackage{mathrsfs}
\usepackage[utf8]{inputenc}
\usepackage[caption=false,font=footnotesize]{subfig}

\usepackage{graphicx}
\usepackage{pifont}
\usepackage{soul}

\newcounter{RNum}
\renewcommand{\theRNum}{\arabic{RNum}}
\newcommand{\Remark}{\noindent\textit{\textbf{Remark}~\refstepcounter{RNum}\textbf{\theRNum}: }}
\newcommand{\NoOne}[1]{\textcolor{red}{#1}}
\newcommand{\NoTwo}[1]{\textcolor{green}{#1}}
\newcommand{\NoThree}[1]{\textcolor{blue}{#1}}
\soulregister\NoOne7
\soulregister\NoTwo7
\soulregister\NoThree7
\soulregister\Remark7

\title{\LARGE \bf
Mutation Sensitive Correlation Filter for Real-Time UAV Tracking with Adaptive Hybrid Label
}

\author{Guangze Zheng, Changhong Fu$^{*}$, Junjie Ye, Fuling Lin, and Fangqiang Ding
\thanks{*Corresponding Author}
\thanks{The authors are with the School of Mechanical Engineering, Tongji University, 201804 Shanghai, China.
        {\tt\small changhongfu@tongji.edu.cn}}%
}

\begin{document}

\maketitle
\thispagestyle{empty}
\pagestyle{empty}

\begin{abstract}
	Unmanned aerial vehicle (UAV) based visual tracking has been confronted with numerous challenges, \textit{e.g.}, object motion and occlusion. These challenges generally introduce unexpected mutations of target appearance and result in tracking failure. However, prevalent discriminative correlation filter (DCF) based trackers are insensitive to target mutations due to a predefined label, which concentrates on merely the centre of the training region. Meanwhile, appearance mutations caused by occlusion or similar objects usually lead to the inevitable learning of wrong information.
	To cope with appearance mutations, this paper proposes a novel DCF-based method to enhance the sensitivity and resistance to mutations with an adaptive hybrid label,~\textit{i.e.}, MSCF. The ideal label is optimized jointly with the correlation filter and remains temporal consistency. 
	Besides, a novel measurement of mutations called mutation threat factor (MTF) is applied to correct the label dynamically.
	Considerable experiments are conducted on widely used UAV benchmarks. The results indicate that the performance of MSCF tracker surpasses other 26 state-of-the-art DCF-based and deep-based trackers. With a real-time speed of $\sim$38 frames/s, the proposed approach is sufficient for UAV tracking commissions.
\end{abstract}

\section{Introduction}
	
	In recent years, visual object tracking has facilitates the applications of UAV, especially in path planning~\cite{8794345}, traffic monitoring~\cite{8377077}, and global pose estimation~\cite{9196606}.
	In the process of UAV object tracking, the target is specified primarily and tracked in the subsequent frames. However, the volatile environment as well as the object motion usually causes severe uncertainties. Besides, the limitation of computing resource and battery life on the aerial platform also impede the development of UAV-based tracking.

	\begin{figure}[h]
		\includegraphics[width=0.46\textwidth]{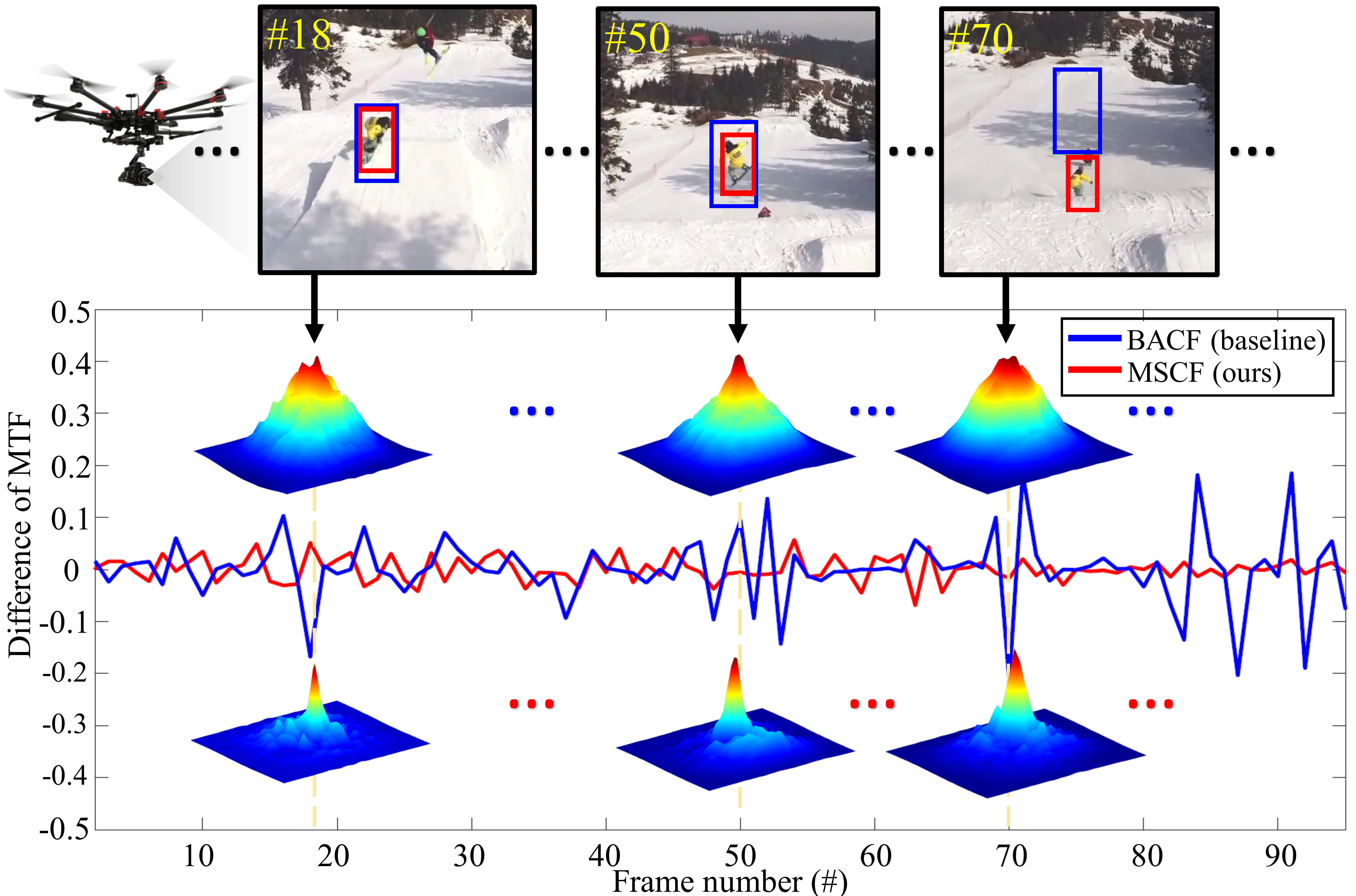}
		\setlength{\abovecaptionskip}{-0.1cm}
		\caption
		{The comparison of tracking performance between the presented MSCF tracker and baseline, BACF tracker~\cite{Galoogahi2017CVPR}. The \color{red} red \color{black} and \color{blue} blue \color{black} boxes are the tracking results from \color{red} MSCF \color{black} tracker and \color{blue} BACF \color{black} tracker respectively. The difference of the proposed MTF metric between frames evaluates the change of mutations, where higher value means drastic transformation. The \color{red} red \color{black} curve with small fluctuations demonstrates that MSCF tracker possesses higher and stronger mutation sensitivity and anti-mutation power than BACF tracker.
		}
\label{fig:Fig_1}
	\end{figure}

	To solve the above problems, DCF-based tracking method has been introduced due to high computational efficiency and promising accuracy~\cite{Fu2020Correlation}. Since the proposition of KCF in~\cite{Henriques2015TPAMI}, meaningful improvements have been made by follow-up researchers~\cite{IBRI, Lin2020TCSVT, fu2021learning}.
	The general pipeline of DCF-based UAV tracking is summarized as follows. The current feature model is updated by the linear interpolation of previous one and current extracted feature with a learning rate. Subsequently, through the ridge regression, a filter is trained by minimizing the squared error of the correlation response with the current feature model. As for position estimation, a new frame is transmitted to UAV and processed with the filter to detect the object for continuous tracking.
	Since the computation is transformed into the frequency domain through fast discrete Fourier transform (DFT), the computation speed is considerably improved. 
	
	Although DCF-based UAV tracking has gained excellent performance, there are still some problems caused by tracking mutations. 
	First, the frequently-used Gaussian label pays most of attention to the centre of training region, but the object generally moves to the neighborhood in case of fast motion.
	Meanwhile, since target mutations, such as similar objects, are manifested as sub-peaks in the response map, they commonly represent severe threats to the detection of the target. Therefore, how to construct an adaptive label and improve the robustness of the tracker against the mutations 
	has a become a crucial issue.

	\begin{figure*}[!t]	
	\centering
	\includegraphics[width=0.85\linewidth]{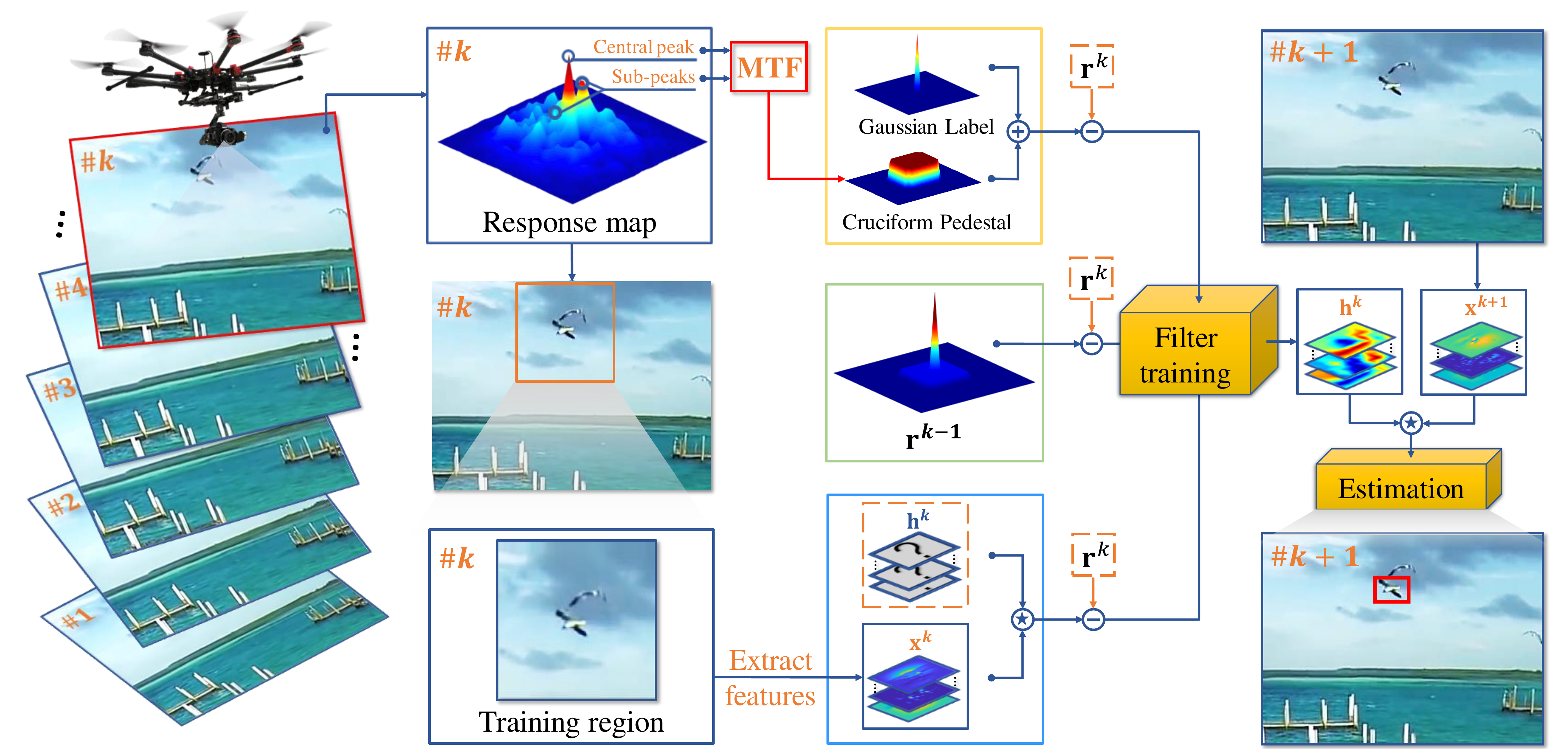}
	\caption
	{Tracking procedure of the proposed MSCF tracker. Dashed boxes denote the variables to be solved in the main regression. As MTF in the \color{red} \textbf{red} \color{black} box is generated from search region in frame ${k}$, it is applied to adjust the altitude value of the cruciform pedestal. Besides, signal ${-}$ denotes the minus operation in the objective function, and ${\star}$ denotes circular correlation. 
	}  
	\label{fig:main}
	\vspace{-0.4cm}
\end{figure*}

	Based on the analysis above, this work proposes a UAV tracking approach with a mutation sensitive correlation filter (MSCF). Main work-flow of MSCF adopts the idea of genetic algorithm, which includes selection, crossover, and evolution. Response maps and peaks in them are regarded as chromosomes and genes respectively. In this way, the mutations can be detected through the proposed MTF measurement and the dangerous threats are consequently repressed.
	During the crossover phase, three labels, \textit{i.e.}, ideal hybrid label, previous label, and current correlation value label, are hybridized to form an adaptive label. The filter evolves afterwards for object detection in the next frame. A qualitative comparison between MSCF and its baseline BACF~\cite{Galoogahi2017CVPR} is shown in Fig.~\ref{fig:Fig_1}, where the tracking object is undergoing drastic deformation in appearance, and mutations are reflected as sub-peaks in response maps. The tracking procedure of MSCF is illustrated in Fig.~\ref{fig:main}. Main contributions of this work can be summarized as follows:
	
	\begin{itemize}
		\item A novel assessment measure of response maps is proposed, namely mutation threat factor (MTF), and the idea of genetic algorithm is adopted.
		\item A novel adaptive hybrid label with MTF adjustment is applied for the filter training. 
		\item A novel MSCF tracker is developed to cope with tracking mutations by employing the adaptivity of the adaptive hybrid label. Numerous experiments are conducted on extensively-used UAV benchmarks. The results demonstrate the superior performance of the MSCF tracker against other 26 state-of-the-art trackers.
	\end{itemize}

	\section{Related Work}
	
	\subsection{DCF for UAV Tracking}
	Since a prior work~\cite{Bolme2010CVPR} introduced correlation filter into visual tracking, methods based on correlation filter have raised a paradigm shift. Subsequently, J. F. Henriques~\textit{et al}.~\cite{Henriques2015TPAMI} contributed a basic framework to DCF-based methods.
	Employments of scale estimation~\cite{Danelljan2017PAMI} and real background information~\cite{Galoogahi2017CVPR} have also brought about further progress of DCF's performance. Meanwhile, efficient feature extraction~\cite{Danelljan2014CVPR, Danelljan2016ECCV} contributes to more precise and comprehensive object representation.
	Nevertheless, standard DCFs commonly use a predefined Gaussian label for regression. These trackers are still weak to handle tracking mutations and prone to collapse as the wrong information is learned. 
	\subsection{Adaptive Label for DCF}
	The mutation-adaptive potential is significant for UAV tracking, especially in extreme situations from aerial views. Nevertheless, the attention paid to 
	adaptive label learning was insufficient. A. Bibi \textit{et al}.~\cite{2016Target} proposed a generic framework which adaptively changed the labels from frame to frame. R. Han \textit{et al}.~\cite{2020Fast} improved the the label adaptivity by assigning different weights to the training samples. 
	However, the potential of the adaptive label has not been achieved thoroughly since the optimization of labels are still lack of direction.
	The deficiency of the label which merely focuses on the object centre is also unsolved due to the inflexibility of the label shape.
	\subsection{Evaluation of Response Map}
	Previous studies have proposed several methods to evaluate the credibility of response maps. D. S. Bolme \textit{et al}. ~\cite{Bolme2010CVPR} took the advantage of the peak to sidelobe ratio (PSR) for failure detection. M. Wang \textit{et al}.~\cite{2017Large} proposed the average peak-to-correlation energy (APCE) to measure the fluctuated degree of response maps. Y. Han \textit{et al}.~\cite{2019State} discovered the application of kurtosis to measure the peakedness and tail weight of the probability distributions of the sampled data. Whereas, certain challenges, \textit{e.g}, similar objects or partial occlusion, generally bring specific mutations which are represented as sub-peaks in the response maps. These methods show insufficient concern for particular dangerous sub-peaks and thus are insensitive to mutations.

\section{Proposed Method}
To achieve higher adaptability to tracking mutations, the genetic idea is adopted in the process of tracking. Labels are regarded as chromosomes while the peaks in the response map are considered as genes. Since dangerous mutations are picked by employing MTF in selection phase, the labels are then hybridized into the adaptive hybrid label in the crossover phase. Finally, the filter and the label are trained in the evolution phase.

\subsection{Selection Phase: Mutation Sensitivity}

\begin{figure}[!t]	
	\includegraphics[width=0.43\textwidth]{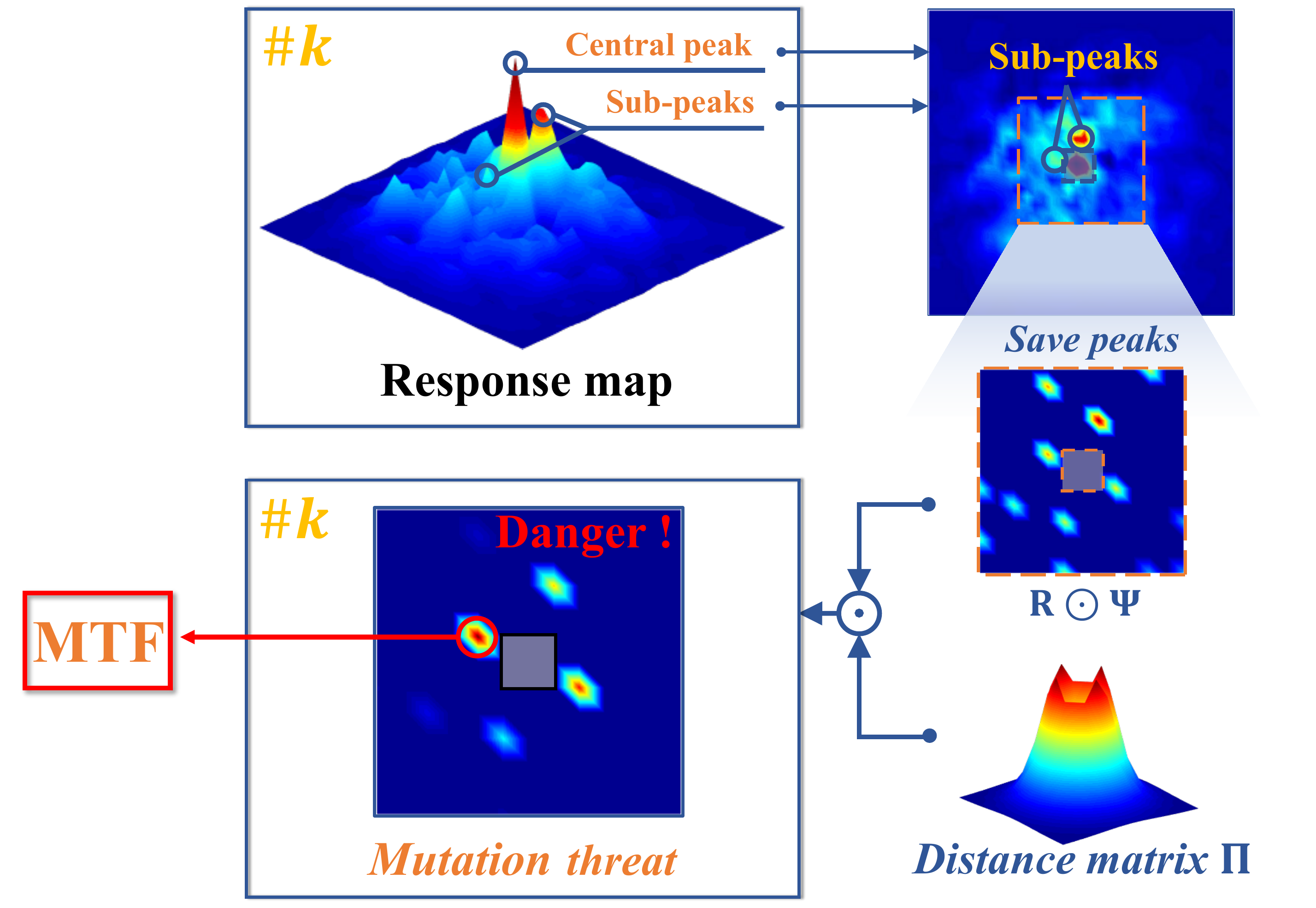}
	\centering
	\setlength{\abovecaptionskip}{-0.1cm}
	\caption
	{
		Main work-flow of MTF generation. 
	}
\label{fig:fig3}
\end{figure}

When the target appearance suffers from mutations, the response maps will fluctuate severely and consequently result in sub-peaks. In this section, the mutation threat factor (MTF) is proposed to represent the threats of sub-peaks in the response map and the process is demonstrated in Fig.~\ref{fig:fig3}, where mutations are demonstrated as sub-peaks in the response map. Through the application of a distance matrix $\mathbf{\Pi}$, MTF is evaluated and utilized in the adaptive hybrid label in selection phase. The principle of selection phase is to find the most dangerous sub-peak to demonstrate the severity of current mutation. During the process of tracking, the threat of sub-peaks in the $k$-th frame is measured as:
\begin{equation}\footnotesize
\mathbf{M}^k=\frac{\mathbf{R}^k\odot\mathbf{\Psi}^k}{{R^k_{max}}}\odot\mathbf{{\Pi}}
\quad,
\end{equation}
where $k$ denotes the $k$-th frame, $\mathbf{R}$ indicates the response map among the search region, and $R^k_{max}$ is the maximum value in $\mathbf{R}$. Besides, $\mathbf{\Psi}$ represents a binary matrix to show the sub-peaks in the response map. MTF in $k$-frame is the maximum value of the calculated $\mathbf{M}^k$. $\mathbf{\Pi}$ is defined as:
\begin{equation}\footnotesize
\mathbf{\Pi}=\left\{
\begin{aligned}
\frac{\nu}{1+\delta\cdot{\exp(d_{(i,j)})}}    &,     &{d_{(i,j)}>d_{min}}\\
0\quad\quad \quad\          &,     &{d_{(i,j)}\le{d_{min}}}\\
\end{aligned}
\quad, \right.
\end{equation}
where $\nu$ denotes the weights in $\Pi$ while $\delta$ describes the influence degree of the distance. Both parameters are defined in advance. 
$d_{(i,j)}$ indicates the distance from the point with $i$-th row and $j$-th column to the centre in the matrix.
Further distance from the centre leads to lower value in $\Pi$. On account that the centre peak of the response map can affect the search of sub-peaks, the weights in the central area are set to 0 with a threshold $d_{min}$.

According to the definition above, the threat from mutations can be expressed elaborately by MTF and the feedback of MTF will be immediately utilized in the adaptive hybrid label. 

\subsection{Crossover Phase: Adaptive Hybrid Label} 
The proposed adaptive hybrid label $\mathbf{r}^k$ in the $k$-th frame is composed of three parts, \textit{i.e}., ideal hybrid label, previous label, as well as the circular correlation value
in the training region. Through the crossover of the three components, the adaptability of label can realize regularization in both temporal and spatial domains.

\textbf{Spatial regularization: ideal hybrid label.} The optimized label consists of two parts: \textbf{(}$\mathbf{i}$\textbf{)} a sharp peak from the Gaussian label; \textbf{(}$\mathbf{ii}$\textbf{)} a cruciform pedestal in the bottom. As for the pedestal altitude value, the generated MTF is introduced to adaptively adjust the value in case of mutations. Meanwhile, the size of the pedestal is initialized according to the target scale and a predefined size ratio parameter. Considering the appearance mutations, the optimal adaptive hybrid label $\mathbf{\Omega}$ is defined as:
\begin{equation}\footnotesize
\mathbf{\Omega}=\mathbf{y}_1+(1-\theta\cdot\max(\mathbf{M}^k))\mathbf{y}_2
\quad,
\end{equation}
\noindent where $\mathbf{y}_1$ and $\mathbf{y}_2$ respectively represent Gaussian label and cruciform pedestal label. $\theta$ is a predefined coefficient, while $\max(\mathbf{M}^k)$ indicates the MTF.

\textbf{Temporal regularization: previous label.} To guarantee the label not to be influenced excessively by particular mutations, the consistency of the hybrid label is updated by referring to the label in the previous frame.
The consistency of the adaptive hybrid label is also taken into account in objective optimization, which is demonstrated as the fourth term in Eq.~(\ref{equ: main}). 

\textbf{Hybridization procedure:}
Different from conventional trackers, correlation filter $\mathbf{h}$ and the hybrid label $\mathbf{r}$ are trained alternately in the proposed method. During the training process in the $k$-th frame, the interaction of the filter and hybrid label contributes to a more rapid convergence. Therefore, the filter and the label have formed a mutual constraint relationship, which promotes the robustness of the tracker.
\subsection{Evolution Phase: Objective Function}
Based on the aforementioned two phases, the adaptability to mutations of the filter and the label can be realized through the evolution phase. Both the filter and the label evolve for higher adaptivity to mutations in solving the problem. The objective function of MSCF can be written as:
\begin{equation}\label{equ: main}\footnotesize
\begin{split}
&\mathcal{E}(\mathbf{h}^k,\mathbf{r}^k)
=
\frac{1}{2} \left\Vert \mathbf{r}^k
- \sum_{d=1}^{D} (\mathbf{P} \mathbf{h}^k_d)\star \mathbf{x}^k_d  \right\Vert_{2}^2 \\ 
\ 
&\quad\quad\ \; + \frac{\lambda_1}{2}\sum_{d=1}^{D}\left\Vert \mathbf{h}^k_d\right\Vert_{2}^2+ \frac{\lambda_2}{2}\left\Vert \mathbf{\Omega}^k-\mathbf{r}^k\right\Vert_{2}^2\\
&\quad\quad\ \; + \frac{\phi}{2}\left\Vert\mathbf{r}^k-\mathbf{r}^{k-1}\right\Vert_{2}^2 
\end{split}
\quad,
\end{equation}
where $d$ denotes the $d$-th feature channel, $k$ indicates the $k$-th frame, $\mathbf{x}\in{\mathbb{R}}^T$ denotes the vectorized image, and $\mathbf{h}\in{\mathbb{R}}^T$ indicates the filter. 
Besides, $\mathbf{r}^k$ denotes the adaptive hybrid label.
$\mathbf{P} \in \mathbb{R}^{N\times{T}}$ is a binary matrix which crops the mid $N$ elements of $\mathbf{x}$, where $T\gg{N}$. $T$ denotes the length of $\mathbf{x}$.
In addition, $\lambda_1$, $\lambda_2$, and $\phi$ are regularization parameters. 
After transformed into frequency domain for computational efficiency, Eq.~(\ref{equ: main}) can be expressed as:
\begin{equation}\label{equ: frequency}\footnotesize
\begin{split}
&\hat{\mathcal{E}}(\mathbf{h}^k,\mathbf{r}^k,\hat{\mathbf{g}}^k) 
=
\frac{1}{2T} \left\Vert \hat{\mathbf{r}}^k
-  \hat{\mathbf{X}}^k\hat{\mathbf{g}}^k \right\Vert_{2}^2 +  \frac{\lambda_1}{2}\left\Vert \mathbf{h}^k\right\Vert_{2}^2 \\
\
&\quad\quad\quad\quad  + \frac{\lambda_2}{2T}\left\Vert \hat{\mathbf{\Omega}}^k-\hat{\mathbf{r}}^k\right\Vert_{2}^2 + \frac{\phi}{2T}\left\Vert\hat{\mathbf{r}}^k-\hat{\mathbf{r}}^{k-1}\right\Vert_{2}^2 \\
&\quad\quad\quad s.t. \quad \hat{\mathbf{g}}^k = \sqrt{T} (\mathbf{I}_D\otimes\mathbf{FP}^{\mathsf{T}})\mathbf{h}^k
\end{split}
\quad,
\end{equation}
where $\hat{\mathbf{g}}^k$ is an auxiliary variable vector and $\mathbf{F}\in{\mathbb{C}^{T\times{T}}}$ denotes the orthonormal matrix. $\hat{\mathbf{X}}^k$ is defined as $\hat{\mathbf{X}}^k=[diag(\hat{\mathbf{x}}^k_1)^\mathsf{T},diag(\hat{\mathbf{x}}^k_2)^\mathsf{T},...,diag(\hat{\mathbf{x}}^k_D)^\mathsf{T}]\in{\mathbb{C}^{T\times{DT}}}$. ${\mathbf{h}^k}=[{\mathbf{h}}^{k\mathsf{T}}_1,{\mathbf{h}}^{k\mathsf{T}}_2,...,{\mathbf{h}}^{k\mathsf{T}}_D]^\mathsf{T}$ and ${\hat{\mathbf{g}}^k}=[\hat{\mathbf{g}}^{k\mathsf{T}}_1,\hat{\mathbf{g}}^{k\mathsf{T}}_2,...,\hat{\mathbf{g}}^{k\mathsf{T}}_D]^\mathsf{T}$ are representations of concatenated channels  separately for $\mathbf{h}$ and $\hat{\mathbf{g}}$. Besides, the symbol ${\hat{}}$ denotes the DFT of a signal. $I_D \in \mathbb{R}^{D\times{D}}$ is an identity matrix. The operator $\otimes$ and superscript $\mathsf{T}$ are Kronecker production and conjugate transpose operation respectively. By applying alternating direction method of multipliers (ADMM)~\cite{StephenDistributed}, the augmented Lagrangian form of Eq.~(\ref{equ: frequency}) is formulated as:
\begin{equation}\footnotesize
\begin{split}
\hat{\mathcal{L}}&(\mathbf{h}^k, \hat{\mathbf{g}}^k,\mathbf{r}^k ,\hat{\zeta}^k) 
=
\hat{\mathcal{E}}(\mathbf{h}^k,\mathbf{r}^k,\hat{\mathbf{g}}^k)+ \hat{\mathscr{L}}(\mathbf{h}^k,\hat{\mathbf{g}}^k,\hat{\zeta}^k) \\
\end{split}
\quad.
\label{equ:freq2}
\end{equation}
where $\hat{\mathscr{L}}(\mathbf{h},\hat{\mathbf{g}},\hat{\zeta})=\frac{\mu}{2} \left\Vert \hat{\mathbf{g}} - \sqrt{T}(\mathbf{I}_D\otimes\mathbf{FP}^{\mathsf{T}}) \mathbf{h} +\frac{\hat{\zeta}}{\mu} \right\Vert_{2}^2$ is the augmented Lagrangian item. $\mu$ denotes the penalty factor of the additional item and $\hat{\zeta} = [\hat{\zeta}_1, \hat{\zeta}_2, ..., \hat{\zeta}_D] \in \mathbb{C}^{DT\times1}$ is the Lagrange multiplier. By decomposing Eq.~(\ref{equ:freq2}) into subproblems, ADMM algorithm can be adopted as follows.

\noindent \textbf{Subproblem h:} Given $\hat{\mathbf{g}}^k$, $\mathbf{r}^k$ and $\hat{\zeta}$, the optimal $\mathbf{h}^{k*}$ can be obtained in closed-form by switching back to time domain:
\begin{equation} \label{equ:sub_h}\footnotesize
\begin{split}
&\mathbf{h}^{k*}_{i+1}= \arg \min_{\mathbf{h}^k}\left\{
\begin{array}{ll} \frac{\lambda_1}{2}\left\Vert \mathbf{h}^k\right\Vert_{2}^2 + \hat{\mathscr{L}}(\mathbf{h}^k,\hat{\mathbf{g}}^k,\hat{\zeta}^k)\\
\end{array}\right\}\\
&\quad\ \;=(\frac{\lambda_1}{T}+\mu)^{-1}(\mu\mathbf{g}^k + \zeta^k)
\end{split}
\quad,
\end{equation}
where $i$ denotes the $i$-th iteration.

\noindent \textbf{Subproblem $\hat{\mathbf{g}}$:} Once $\hat{\mathbf{h}}^k$, $\mathbf{r}^k$ and $\hat{\zeta}$ are fixed, the optimization of $\hat{\mathbf{g}}^{k*}$ can be gained as:
\begin{equation} \label{equ:sub_g_1}\footnotesize
\begin{split}
&\hat{\mathbf{g}}^{k*}_{i+1}= \arg \min_{\hat{\mathbf{g}}^k} \left\{ 
\begin{array}{ll}
\frac{1}{2T}\left\Vert \hat{\mathbf{r}}^k- \hat{\mathbf{X}}^k\hat{\mathbf{g}}^k\right\Vert_{2}^2+\hat{\mathscr{L}}(\mathbf{h}^k,\hat{\mathbf{g}}^k,\hat{\zeta}^k)\\
\end{array}\right\}\\
\end{split}
\quad.
\end{equation}

For the sake of lower computation complexity, we sample $\hat{\mathbf{x}}^k$ across all $D$ channels in each pixel as:
\begin{equation} \label{equ:sub_g_2}\footnotesize
\hat{\mathbf{g}}^{k*}(n)= \arg \min_{\hat{\mathbf{g}}^k(n)}\left\{ 
\begin{array}{ll}\vspace{6pt}
\frac{1}{2T}\left\Vert \hat{\mathbf{r}}^k(n)- \hat{\mathbf{x}}^{k}(n)^{\mathsf{T}}\hat{\mathbf{g}}^k(n)\right\Vert_{2}^2\\
+\frac{\mu}{2} \left\Vert \hat{\mathbf{g}}^k(n) - \hat{ \mathbf{h}}^k(n) +\frac{{\zeta}(n)}{\mu} \right\Vert_{2}^2\\
\end{array}\right\}\\
\quad,
\end{equation}
where $\hat{\mathbf{h}}^k(n)=[\hat{\mathbf{h}}_1^k(n),\hat{\mathbf{h}}_2^k(n),...,\hat{\mathbf{h}}_D^k(n)]\in\mathbb{C}^{D\times1}$ and $\hat{\mathbf{h}}^k=\sqrt{D}\mathbf{FP}^{\mathsf{T}}{\mathbf{h}}^k$. Since $\hat{\mathbf{h}}^k$ can be solved efficiently in frequency domain, the solution of $\mathbf{g}^{k*}(n)$ is obtained by:
\begin{equation}\footnotesize
\begin{split}\label{equ:sub_g_3}
&\hat{\mathbf{g}}^{k*}(n) 
=\frac{1}{\mu T} \left(\mathbf{I}_D - \frac{ \hat{\mathbf{x}}^k(n) \hat{\mathbf{x}}^{k{\mathsf{T}}}(n)}{\mu T \mathbf{I}_D + \hat{\mathbf{x}}^{k{\mathsf{T}}}(n) \hat{\mathbf{x}}^k(n)}\right)\rho\\
\end{split}
\quad,
\end{equation}
where the vector $\rho$ is formed as $\rho=\hat{\mathbf{x}}^k(n) \hat{\mathbf{r}}^k+\mu T\hat{\mathbf{h}}^k(n)  - T{\zeta}(n)$. Sherman-Morrison formula~\cite{1950Adjustment} is used in the optimization of Eq.~(\ref{equ:sub_g_3}), \textit{i.e.}, $(\mathbf{A} + \mathbf{u} \mathbf{v}^{\mathsf{T}})^{-1} = \mathbf{A}^{-1} - (\mathbf{I}_D+\mathbf{v}^{\mathsf{T}}\mathbf{A}^{-1}\mathbf{u})^{-1} \mathbf{A}^{-1}\mathbf{u} \mathbf{v}^{\mathsf{T}}\mathbf{A}^{-1}$, where $\mathbf{A}=\mu T\mathbf{I}_D$, and $\mathbf{u} = \mathbf{v} = \hat{\mathbf{x}}^k(n)$. 
In consequence, Eq.~(\ref{equ:sub_g_3}) can be further simplified as:
\begin{equation}\footnotesize
\begin{split}\label{equ:sub_g_4} 
&\hat{\mathbf{g}}^{k*}(n) = \frac{1}{\mu T}( \hat{\mathbf{x}}^k(n) \hat{\mathbf{r}}^k(n)+\mu T\hat{\mathbf{h}}^k(n)  - T{\zeta}(n))\\
& \quad\quad\quad\quad -\frac{\hat{\mathbf{x}}^k(n)}{\mu T b}(\hat{\mathbf{r}}^k(n)\hat{\mathbf{s}}^k_\mathbf{x}(n)-T\hat{\mathbf{s}}^k_\mathbf{\zeta}(n)+\mu T\hat{\mathbf{s}}^k_\mathbf{h}(n))\\
\end{split}
\quad,
\end{equation}
where $\hat{\mathbf{s}}^k_\mathbf{x}(n)= \hat{\mathbf{x}}^{k{\mathsf{T}}}(n)\hat{\mathbf{x}}^k(n)$, $\hat{\mathbf{s}}^k_{\zeta}(n)= \hat{\mathbf{x}}^{k{\mathsf{T}}}(n)\hat{\zeta}$, $\hat{\mathbf{s}}^k_{\mathbf{h}}(n)= \hat{\mathbf{x}}^{k{\mathsf{T}}}(n)\hat{\mathbf{h}}$ and $b=\hat{\mathbf{s}}^k_\mathbf{x}(n)+\mu T$. 

\noindent \textbf{Subproblem r:} The optimal $\hat{\mathbf{r}}^{k*}$ can be obtained in closed-form, on condition that $\hat{\mathbf{h}}^k$, $\mathbf{g}^k$ and $\hat{\zeta}$ are fixed:
\begin{equation} \label{equ:sub_r}\footnotesize
\begin{split}
&\mathbf{r}^{k*}= \arg \min_{\mathbf{r}^k}\left\{
\begin{array}{ll}\vspace{4pt}
\frac{1}{2}\left\Vert \hat{\mathbf{r}}^k- \hat{\mathbf{X}}^k\hat{\mathbf{g}}^k\right\Vert_{2}^2\\\vspace{4pt}
+\frac{(1+\Psi^2)\lambda_2}{2}\left\Vert \hat{\mathbf{\Omega}}^k-\hat{\mathbf{r}}^k\right\Vert_{2}^2 \\
+ \frac{(1-\Psi^2)\phi}{2}\left\Vert\hat{\mathbf{r}}^k-\hat{\mathbf{r}}^{k-1}\right\Vert_{2}^2\\
\end{array}\right\}\\
&\quad\  = \frac{\hat{\mathbf{X}}^k\hat{\mathbf{g}}^k+\lambda_2(1+\Psi^2)\hat{\mathbf{\Omega}}^k+\phi(1-\Psi^2)\hat{\mathbf{r}}^{k-1}}{1+\lambda_2(1+\Psi^2)+\phi(1-\Psi^2)}
\end{split}
\quad.
\end{equation}

\noindent \textbf{Lagrangian multiplier update:} On this occasion, Lagrangian multipliers can be updated as:
\begin{equation}\label{equ:Lag_up}\footnotesize
\hat{\zeta}^{k*}_{i+1} = \hat{\mathbf{\zeta}}^{k*}_{i} + \mu_{i+1} (\hat{\mathbf{g}}^{k*}_{i+1} - \hat{\mathbf{h}}^{k*}_{i+1})
\quad.
\end{equation}
The step size regularization constant $\mu_{i+1}$ is updated as $\mu_{i+1}=min(\mu_{max},\beta\mu_{i})$, where $\beta$ is defined in advance.

\subsection{Object Localization}
As the feature in the searching region of a new frame is extracted, the object can be localized according to the response map $\mathbf{R}^k$, which is generated as:
\begin{equation}\footnotesize
\mathbf{R}^k=\mathcal{F}^{-1}\sum_{d=1}^{D} \hat{\mathbf{z}}_d^k \odot \hat{\mathbf{g}}_d^{k-1}
\quad,
\end{equation}
where $\mathcal{F}^{-1}$ indicates the inverse discrete Fourier transform (IDFT) operator and $\hat{\mathbf{z}}_d^k$ is the extracted feature map of frequency domain in the $k$-th frame. 

\begin{figure*}[!t]
	\centering	
	\subfloat
	{
		\includegraphics[width=0.3\linewidth]{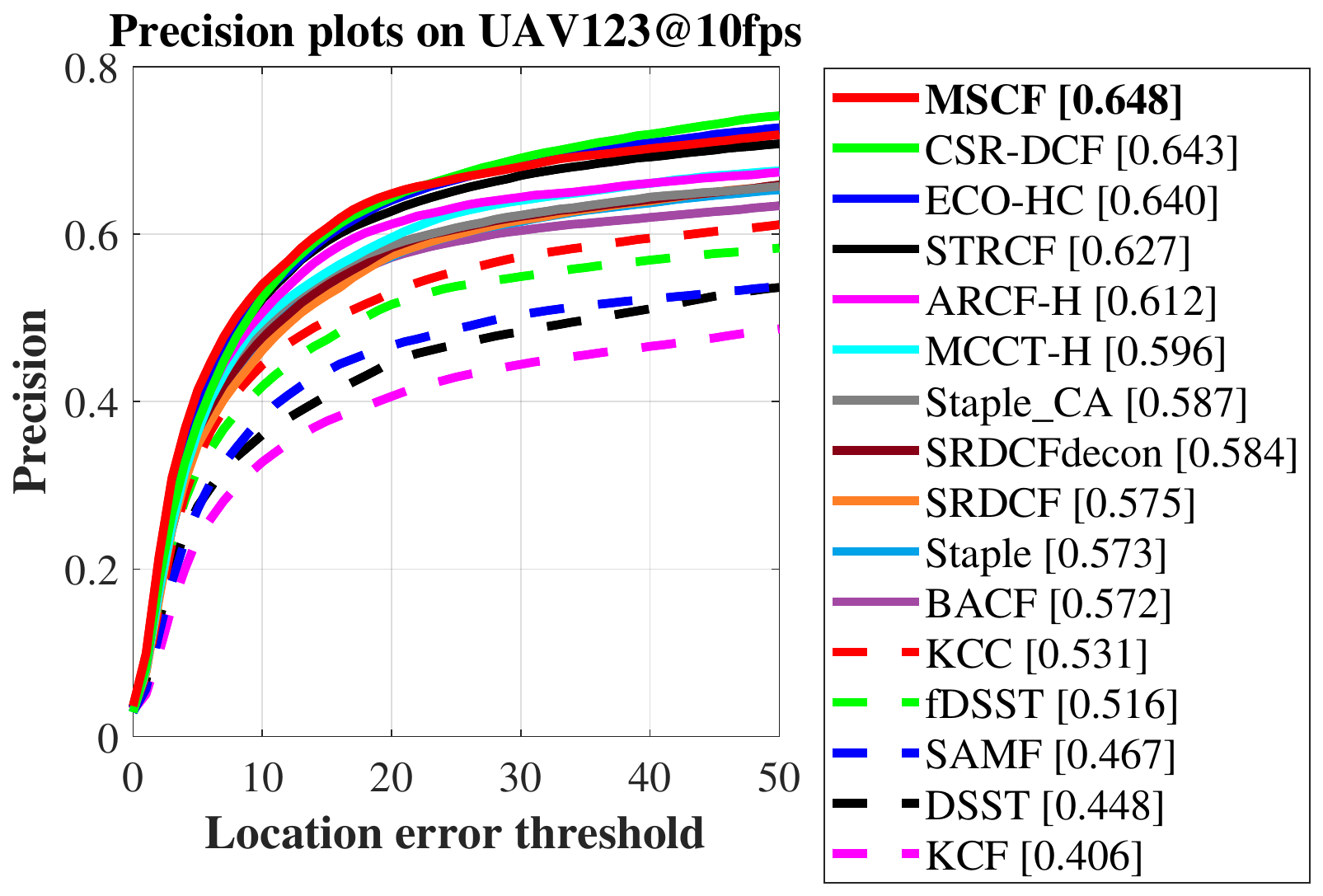}
	}
	\subfloat
	{
		\includegraphics[width=0.3\linewidth]{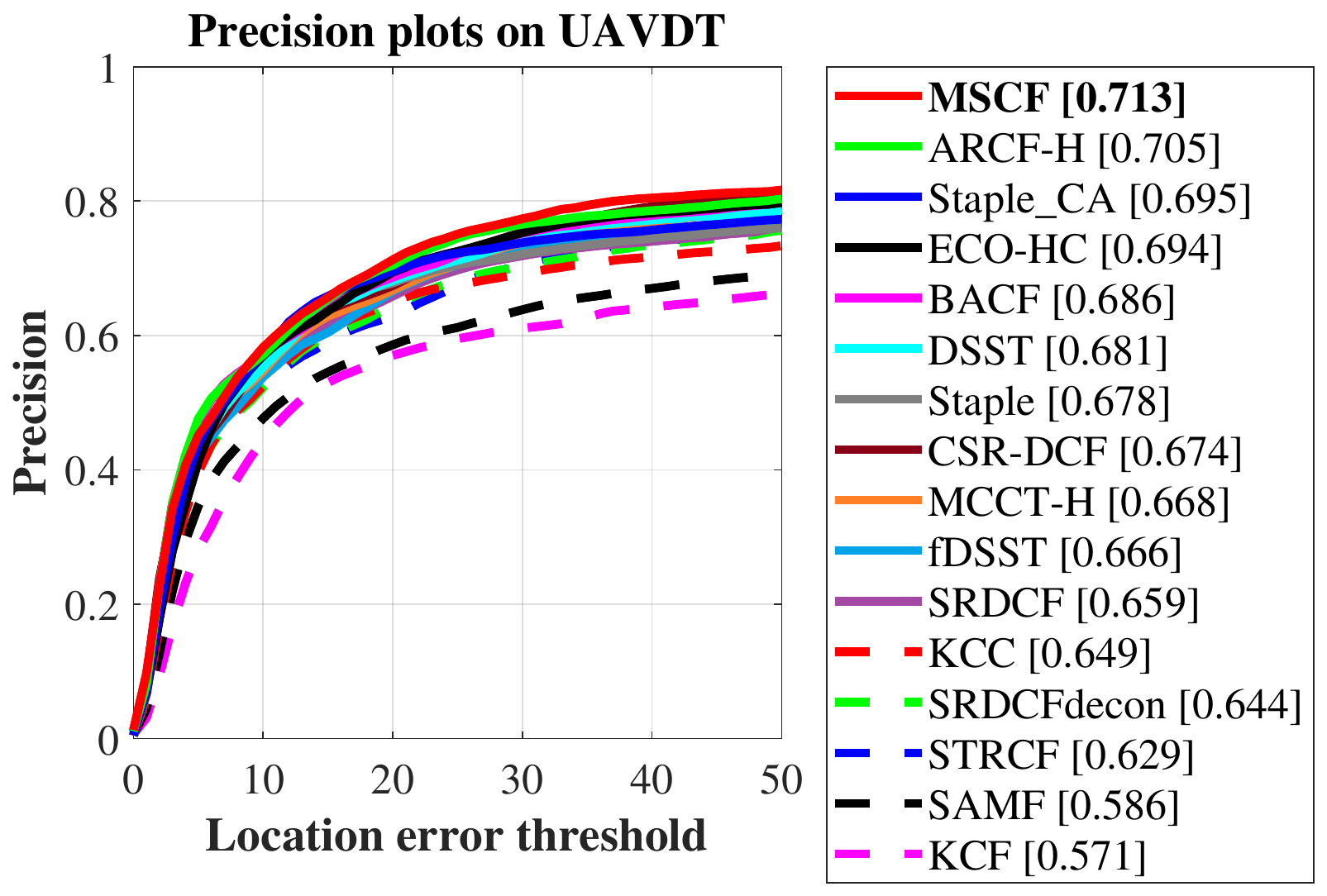}
	}
	\subfloat
	{
		\includegraphics[width=0.3\linewidth]{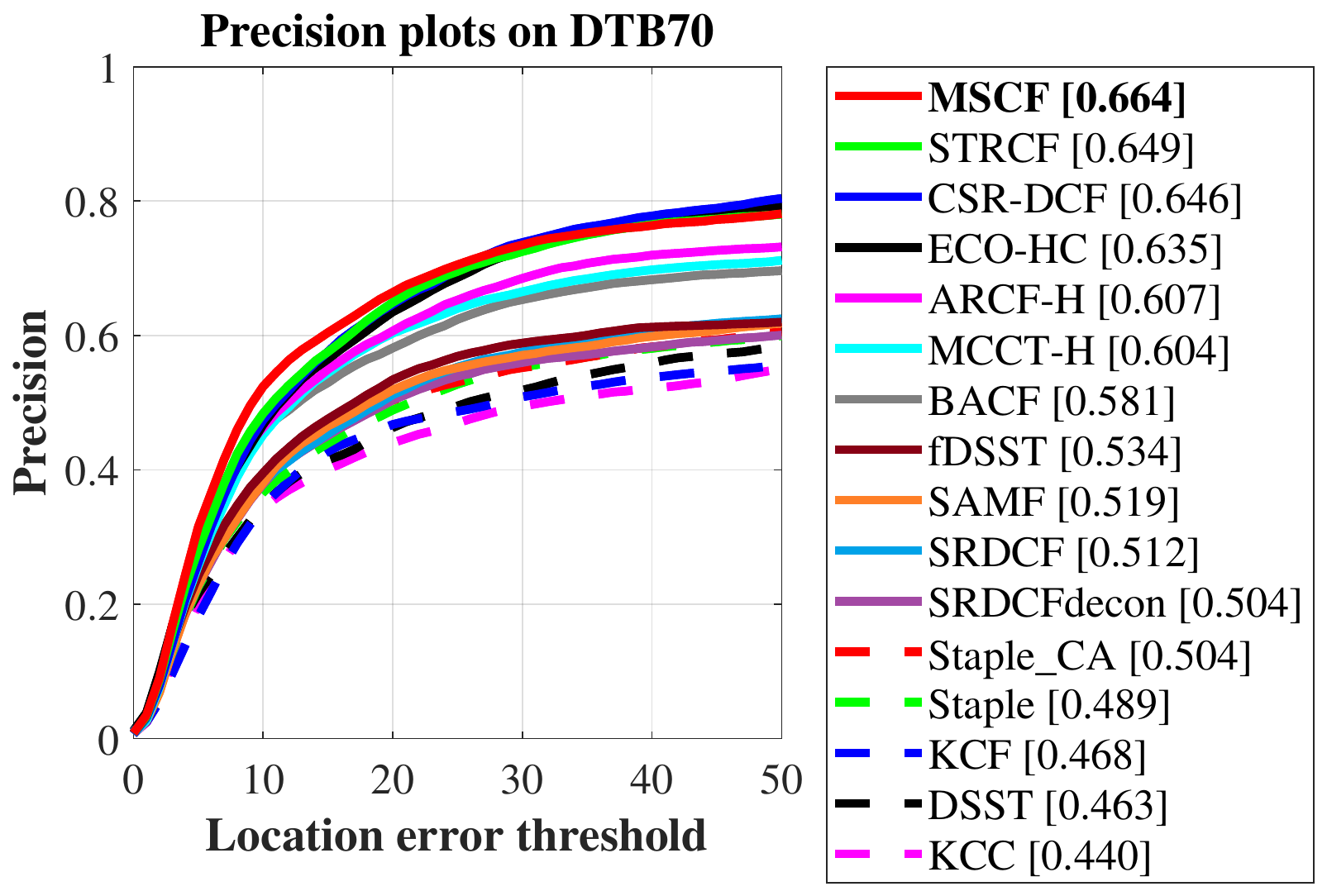}
	}
	\\[-20pt]
\end{figure*}

\begin{figure*}[!h]
	\centering	
	\subfloat[Results on UAV123@10fps.]
	{
		\includegraphics[width=0.3\linewidth]{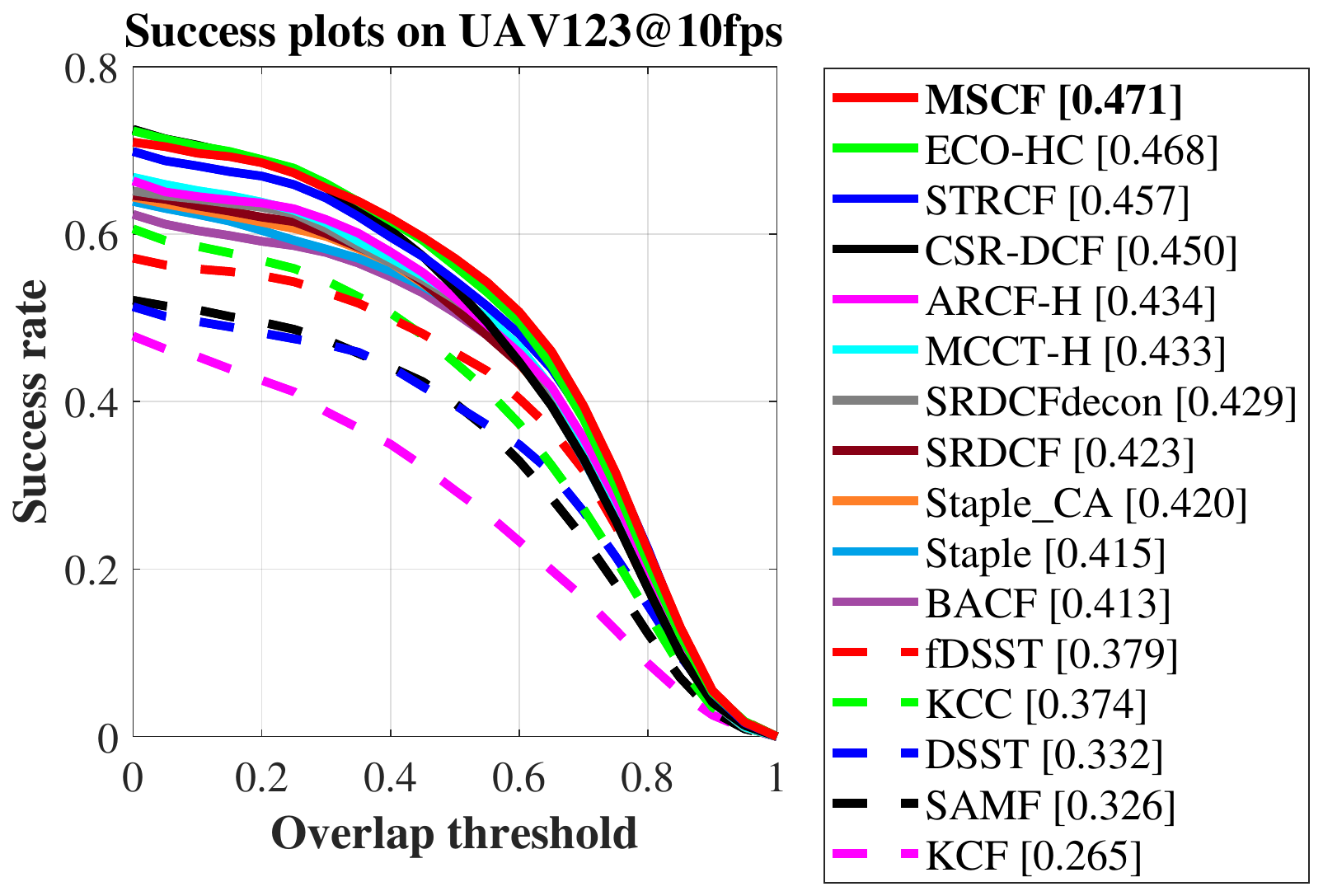}
		\label{fig:UAV123}
	}
	\subfloat[Results on DTB70.]
	{
		\includegraphics[width=0.3\linewidth]{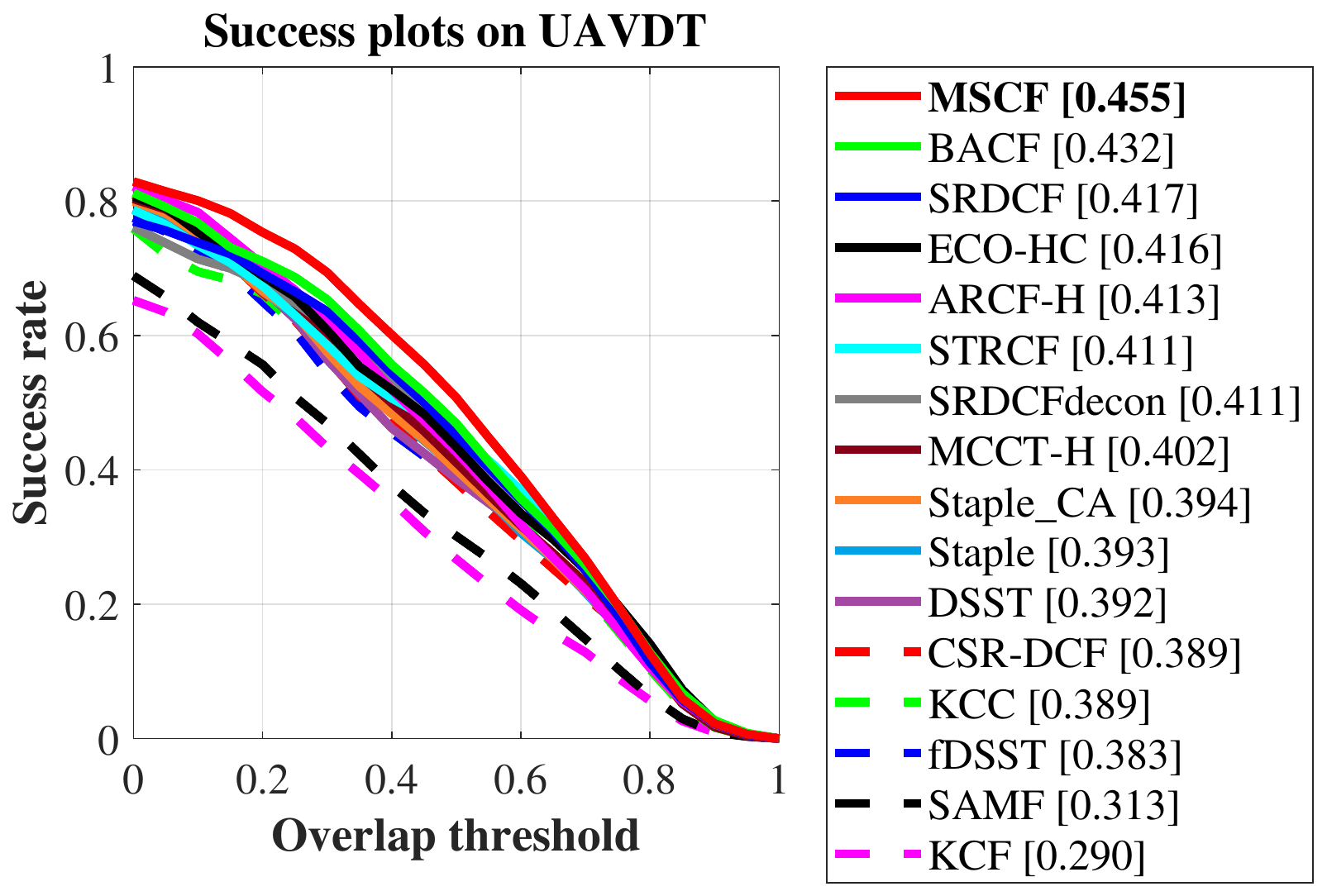}
		\label{fig:DTB70}
	}
	\subfloat[Results on UAVDT.]
	{
		\includegraphics[width=0.3\linewidth]{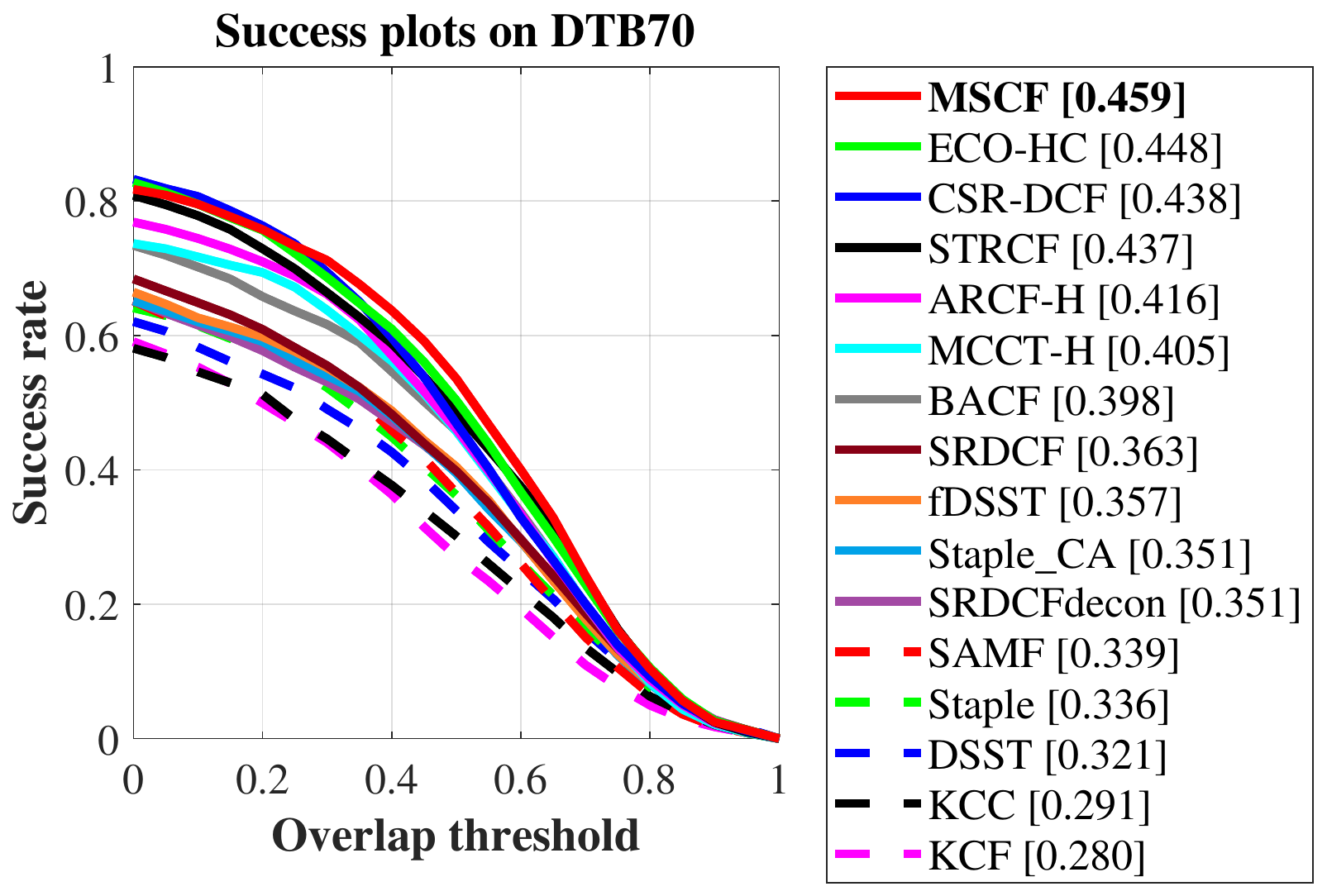}
		\label{fig:UAVDT}
	}
	\caption
	{
		Overall performance evaluation. PPs and SPs of the proposed MSCF tracker and other  15 state-of-the-art trackers. MSCF yields competitive performance on all three benchmarks in both terms of precision and AUC.
	}
	\label{fig:hand_over}
\end{figure*}
\begin{table*}[htbp]
	\centering
	\caption{Overall average performance evaluation. Average precision, success rate and FPS of top 12 trackers on three benchmarks. The best three performances are respectively highlighted with \textbf{\textcolor{red}{red}}, \textbf{\textcolor{green}{green}}, and \textbf{\textcolor{blue}{blue}} colors.}
			\vspace{-10pt}
	\resizebox{\linewidth}{!}{
		\begin{tabular}{lccccccccccc|c}
			\toprule
			Trackers & BACF   & Staple\_CA & SRDCF  & SRDCFdecon & MCCT-H & CSR-DCF & STRCF  & ECO-HC & fDSST  & Staple & ARCF-H & \bf{MSCF} \\
			\midrule
			Prec.  & 0.613  & 0.595  & 0.582  & 0.577  & 0.622  & \textcolor[rgb]{ 0,  0,  1}{\textbf{0.655 }} & 0.635  & \textcolor[rgb]{ 0,  1,  0}{\textbf{0.657 }} & 0.572  & 0.580  & 0.641  & \textcolor[rgb]{ 1,  0,  0}{\textbf{0.675 }} \\
			Succ.  & 0.414  & 0.388  & 0.401  & 0.397  & 0.414  & 0.426  & \textcolor[rgb]{ 0,  0,  1}{\textbf{0.435 }} & \textcolor[rgb]{ 0,  1,  0}{\textbf{0.444 }} & 0.373  & 0.382  & 0.421  & \textcolor[rgb]{ 1,  0,  0}{\textbf{0.462 }} \\
			FPS    & 43.8   & 52.1   & 10.9   & 5.2    & 49.1   & 11.3   & 24.3   & \textcolor[rgb]{ 0,  0,  1}{\textbf{60.8 }} & \textcolor[rgb]{ 1,  0,  0}{\textbf{173.8 }} & \textcolor[rgb]{ 0,  1,  0}{\textbf{91.8 }} & 41.6   & 37.6  \\
			\bottomrule
	\end{tabular}}
	\label{tab:hand_avg}%
\end{table*}%

\section{Experiment}
	In this section, the evaluation experiments of MSCF tracker are conducted on 243 challenging image sequences and altogether over 90,000 frames from three widely applied UAV benchmarks, namely UAV123@10fps~\cite{Mueller2016ECCV}, DTB70~\cite{Li2017AAAI}, and UAVDT~\cite{Du2018ECCV}. Performance of MSCF tracker is compared with other 15 CPU-based trackers, including KCF~\cite{Henriques2015TPAMI}, SAMF~\cite{Li2014ECCVws}, SAMF\_CA~\cite{Li2014ECCVws}, DSST~\cite{Danelljan2014BMVC}, fDSST~\cite{Danelljan2017PAMI}, Staple\_CA~\cite{Bertinetto2016CVPR}, KCC~\cite{Wang2018AAAI}, BACF~\cite{Galoogahi2017CVPR}, CSR-DCF~\cite{Lukezic2017CVPR}, SRDCF~\cite{Ma2015ICCV}, SRDCFdecon~\cite{Ma2015ICCV}, STRCF~\cite{Li2018CVPR}, ECO-HC~\cite{Danelljan2017CVPR}, MCCT-H~\cite{Wang2018CVPR}, ARCF-H~\cite{Huang2019ICCV}, and 11 GPU-based trackers, \textit{i.e.}, DeepSTRCF~\cite{Li2018CVPR}, CoKCF~\cite{zhang2017PR},  ECO~\cite{Danelljan2017CVPR}, CCOT~\cite{Danelljan2016ECCV}, MCPF~\cite{Zhang_2017_CVPR},  UDT~\cite{Wang_2019_CVPR}, UDT+~\cite{Wang_2019_CVPR}, IBCCF~\cite{Li_2017_ICCV}, ASRCF~\cite{Dai2019CVPR}, MCCT~\cite{Wang2018CVPR}, and TADT~\cite{Li_2019_CVPR}.




\subsection{Implementation Details}
For fairness in evaluation, the parameters of the testes trackers are from their official versions, and the tests are conducted on the same platform.
\subsubsection{Parameter setting} 
In terms of ADMM algorithm, we use three-step iterations for optimization. Other parameters are set as: $\mu = 0.1$, $\mu_{max} = 10000$, $\beta = 10$, $\gamma = 27$ and $\gamma_{max}=10000$. The learning rate is set to $0.0158$. The training process is performed every $2$ frames in practice for boosting efficiency. In addition, the regularization parameters are set as: $\lambda_1=20$, $\lambda_2=840$, and $\phi=1$. The hyper parameters for adaptive hybrid label represent as: $\delta=0.01$, $\nu=1$, $\eta=0.044$ and the ratio of pedestal size is set to $2.5$.
 All parameters remain fixed for all image sequences on three benchmarks. Source code is available here:  \url{https://github.com/vision4robotics/MSCF-tracker}.

\subsubsection{Features} MSCF tracker adopts a combination of histogram of oriented gradient (HOG)~\cite{Felzenszwalb2010TPAMI}, color names (CN)~\cite{Danelljan2014CVPR} and gray-scale features for object representation.

\subsubsection{Platform} All the experiments, including the test of the MSCF tracker and others, are conducted in MATLAB R2019a on a PC equipped with an Intel Core i7-8700K CPU (3.70GHz) and an Nvidia GeForce RTX 2080 GPU.

\subsubsection{Evaluation metrics} All evaluations are based on the one-pass evaluation (OPE), which includes two metrics, \textit{i.e.}, precision and success rate~\cite{Wu2015PAMI}. The precision is measured by the center location error (CLE) between the estimated and the ground truth  bounding boxes. In practice, the percentage of frames whose CLE below 20 pixels (as commonly set) in the precision plot (PP) is used to rank the trackers. As for the success rate, the intersection over union (IoU) between the ground truth and estimated results is reported in the success plot (SP), as usual, this work uses the area under the curve (AUC) of SP to rank the trackers. Besides, frames/s is used to rank the tracking speed of the approaches.

\begin{figure*}[!t]
	\centering	
	\subfloat[Fast motion on UAV123@10fps.]
	{
		\includegraphics[width=0.24\linewidth]{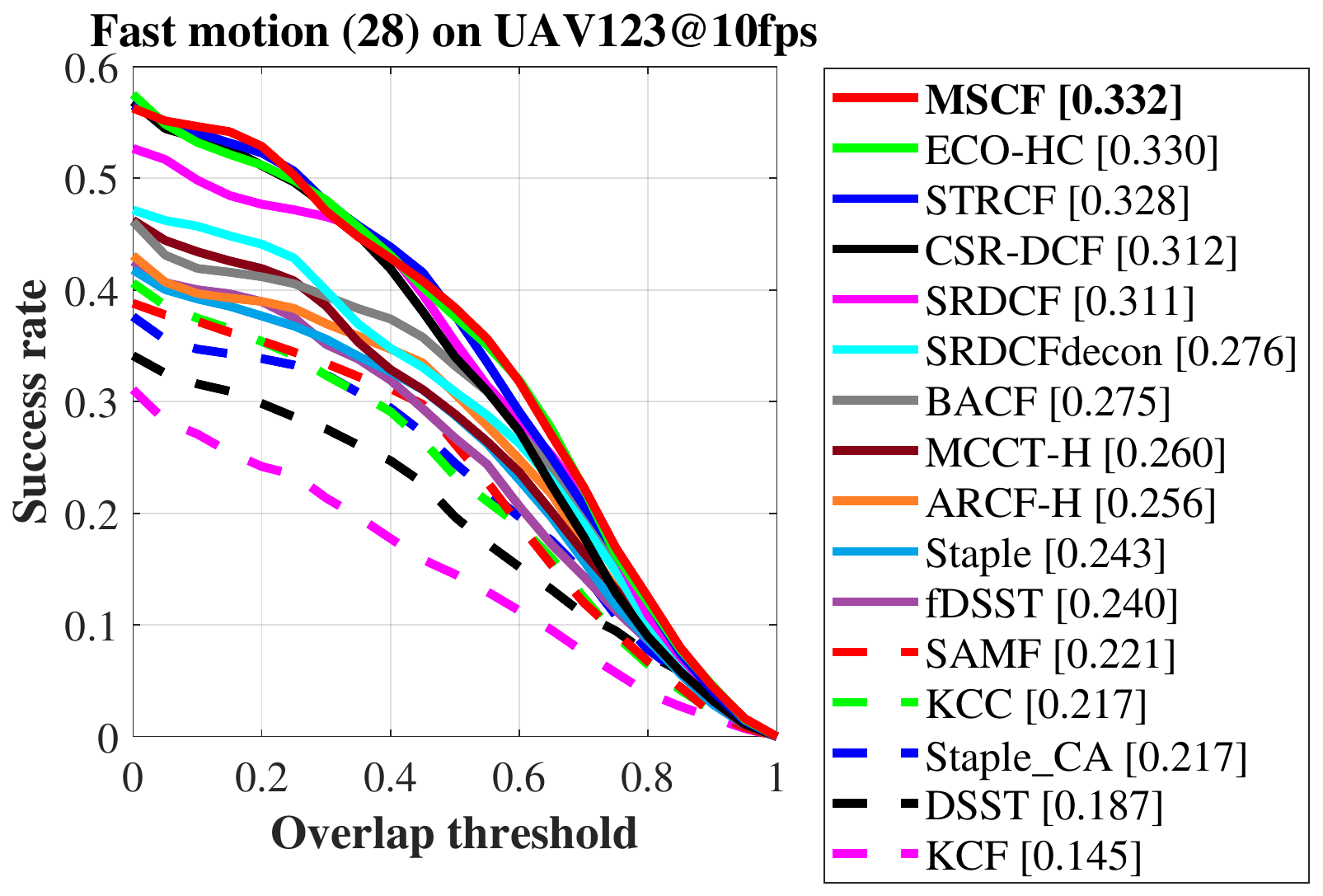}
	}
	\subfloat[Camera motion on UAVDT.]
	{
		\includegraphics[width=0.24\linewidth]{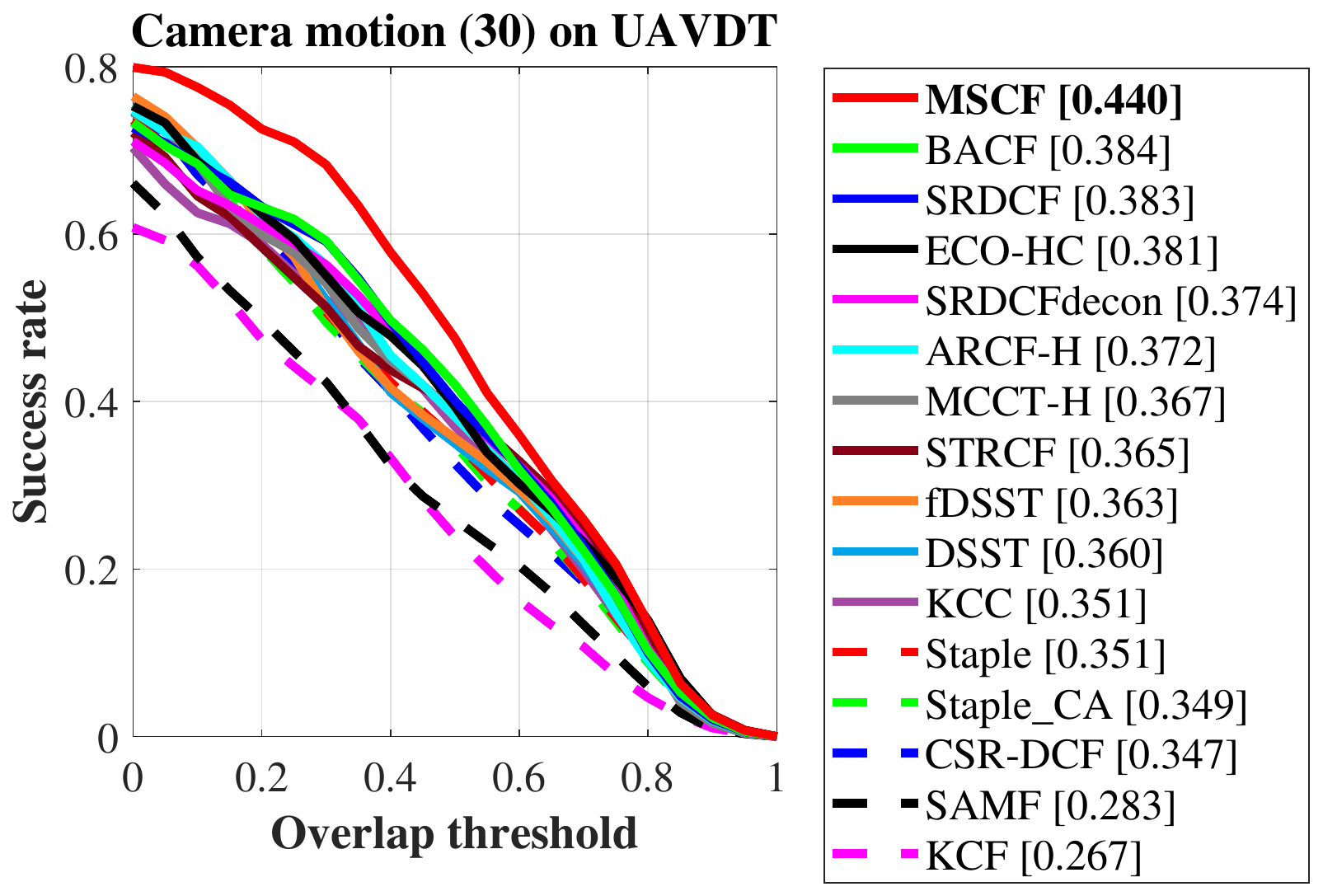}
	}
	\subfloat[Object motion on UAVDT.]
	{
		\includegraphics[width=0.24\linewidth]{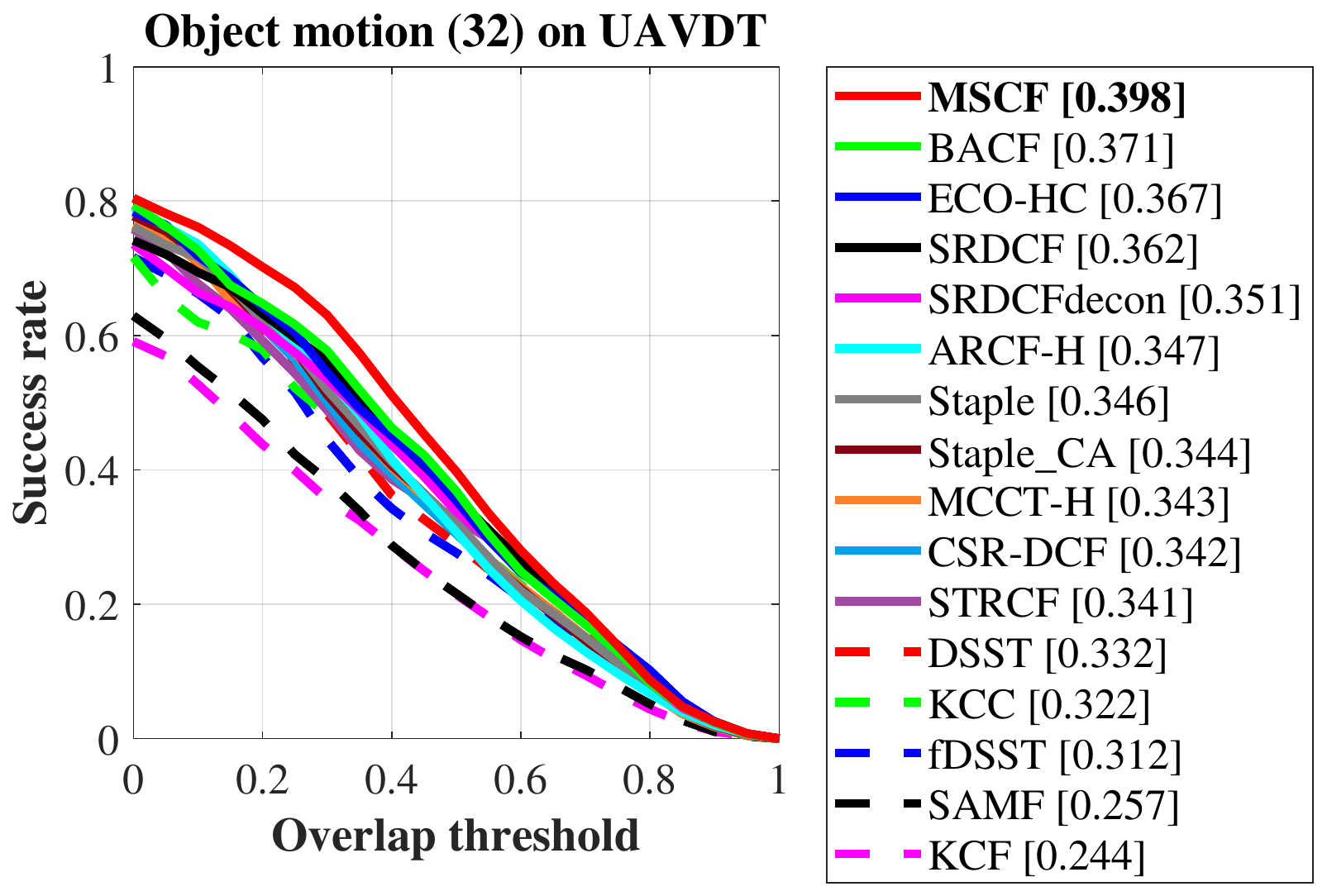}
	}
	\subfloat[Fast camera motion on DTB70.]
	{
		\includegraphics[width=0.24\linewidth]{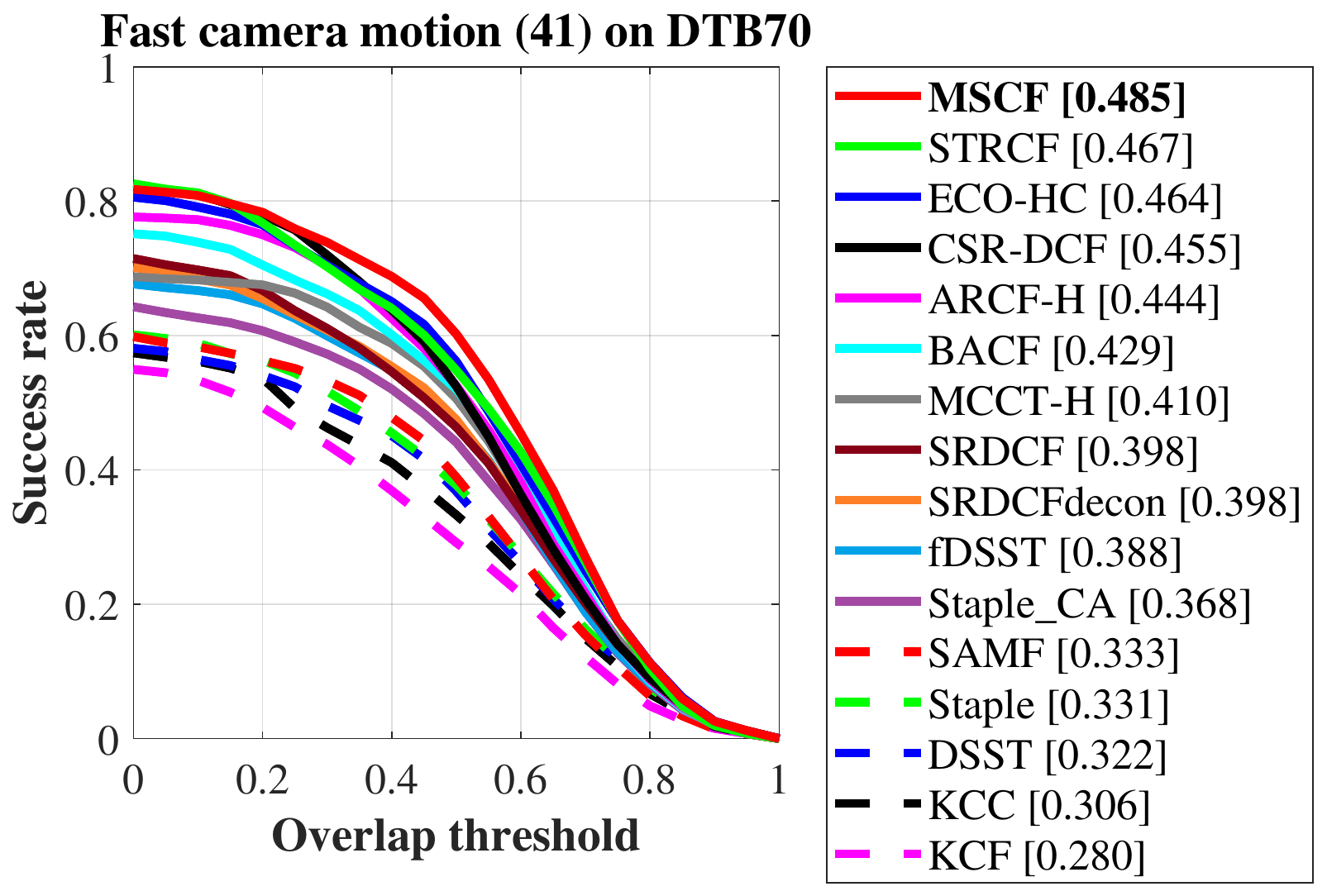}
	}
	\\[-20pt]
\end{figure*}
\begin{figure*}[!h]
	\centering	
	\subfloat[Scale variation on UAV123@10fps.]
	{
		\includegraphics[width=0.24\linewidth]{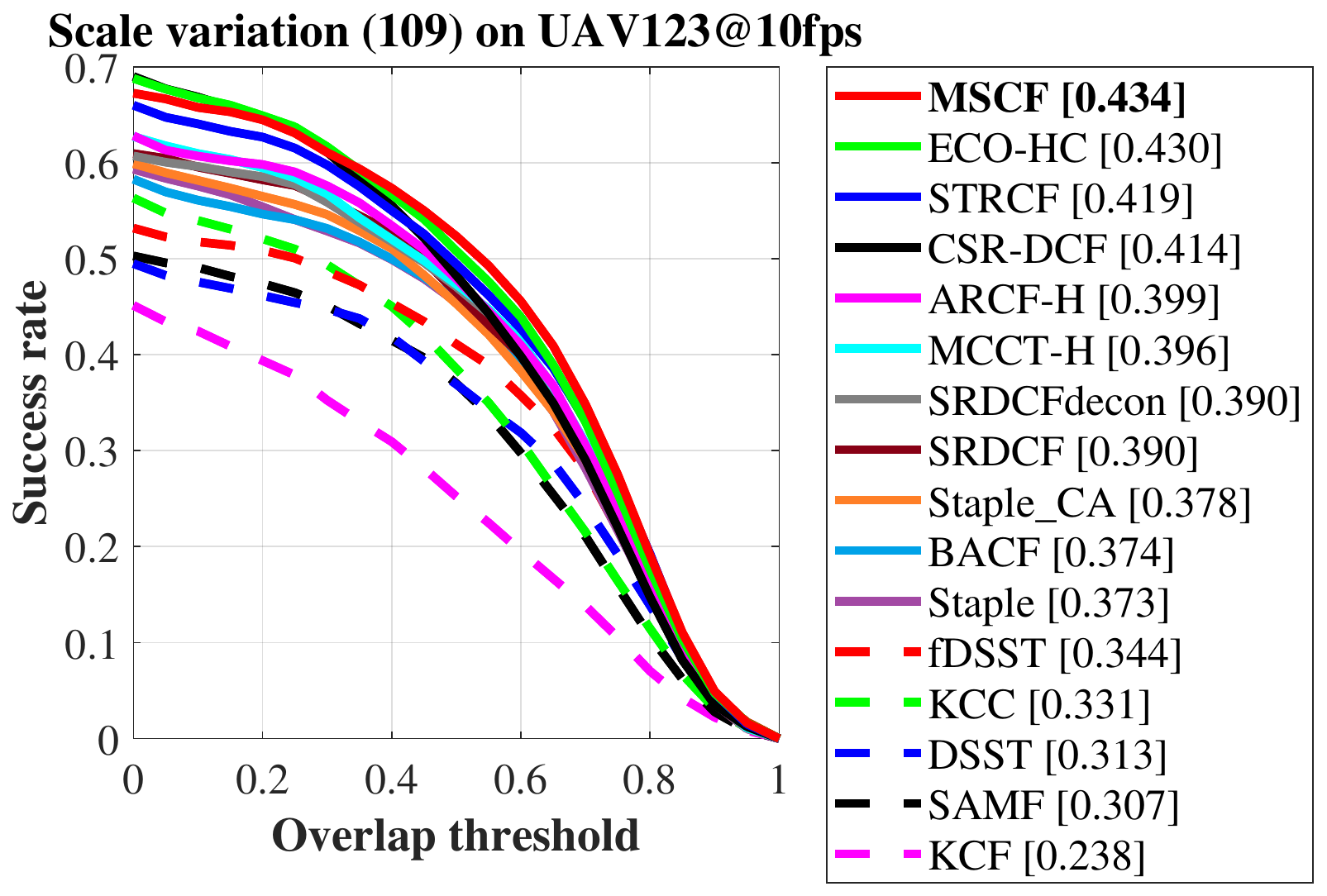}
		\label{fig:UAV123}
	}
	\subfloat[Large occlusion on UAVDT.]
	{
		\includegraphics[width=0.24\linewidth]{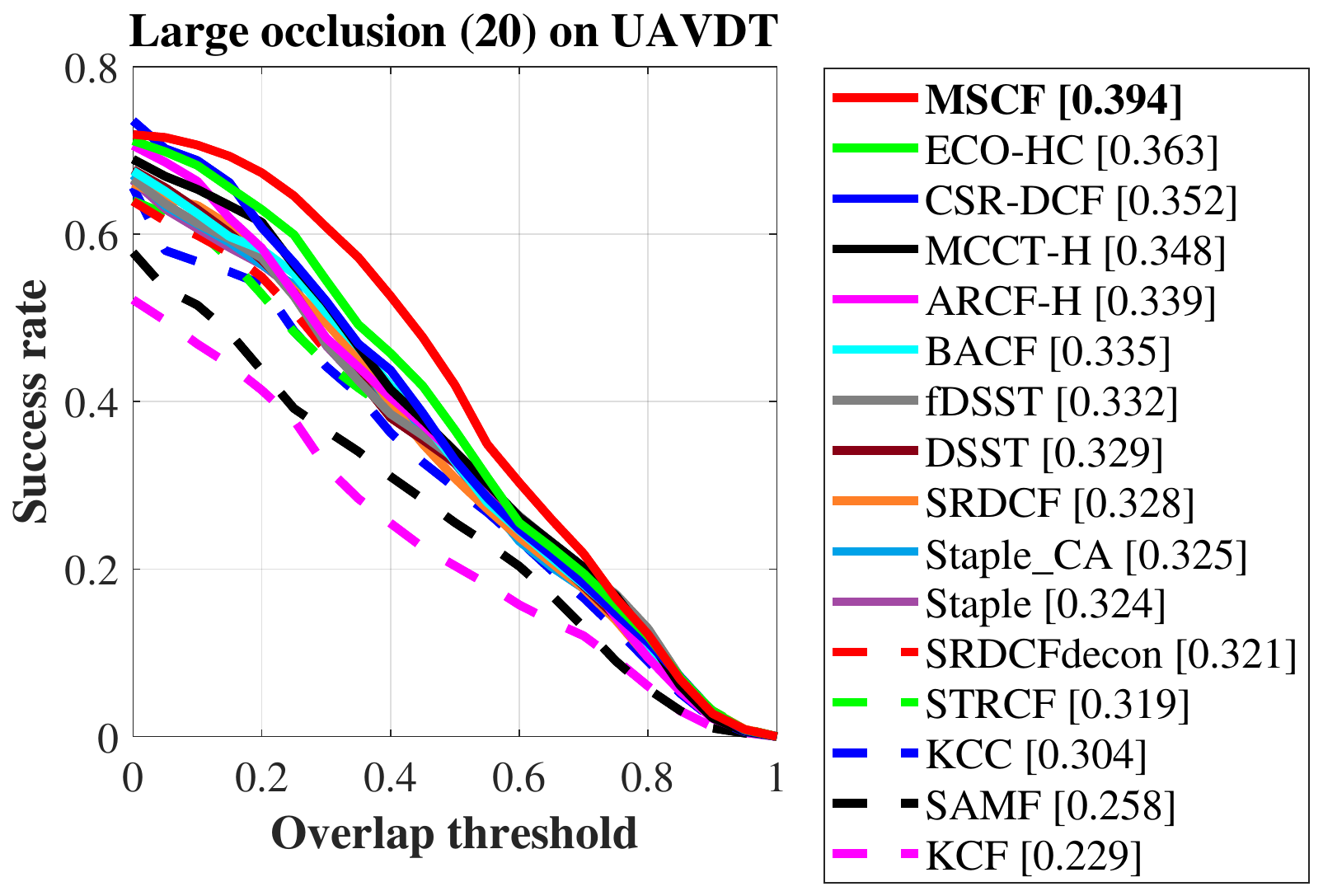}
		\label{fig:DTB70}
	}
	\subfloat[Deformation on DTB70.]
	{
		\includegraphics[width=0.24\linewidth]{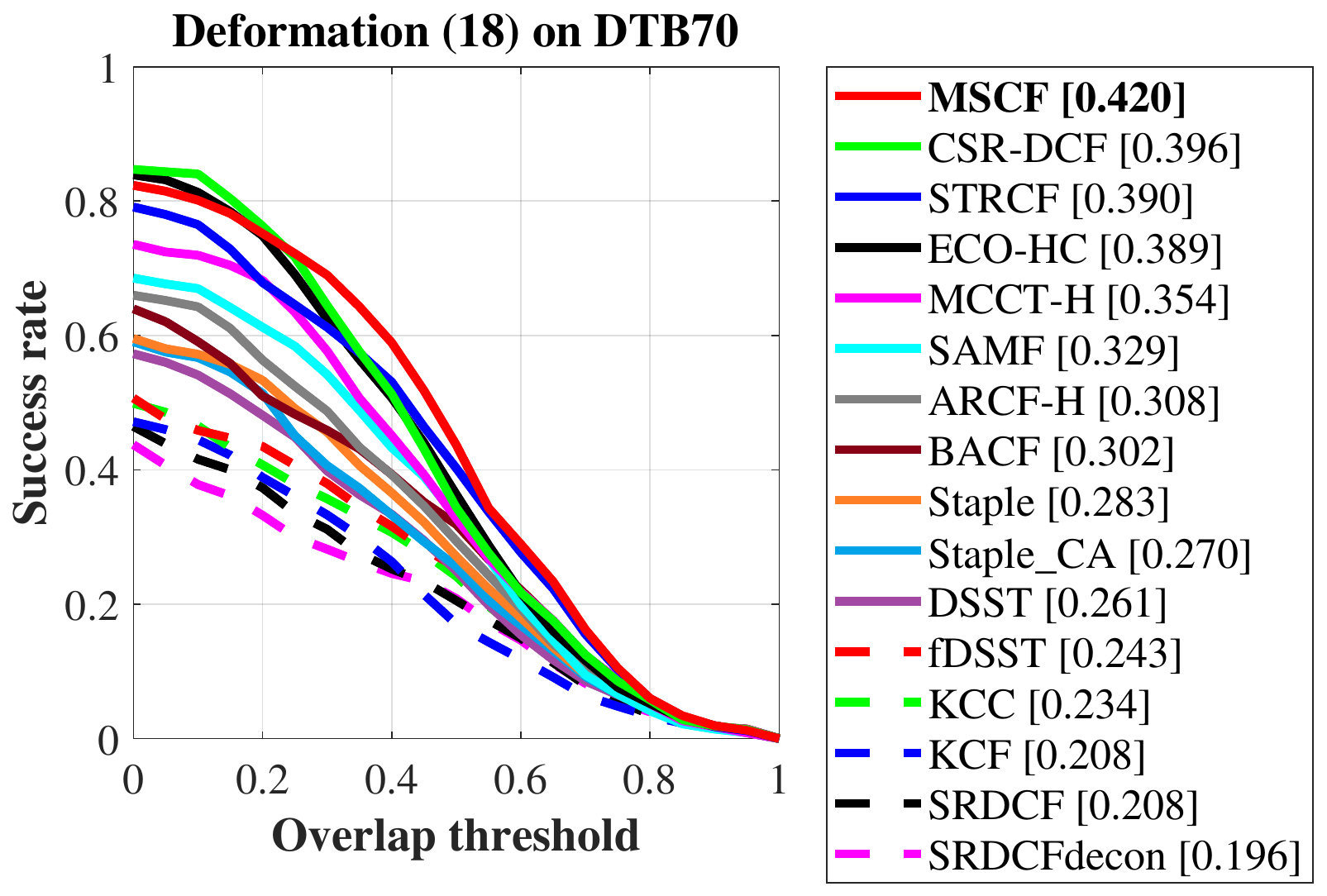}
		\label{fig:UAVDT}
	}
	\subfloat[Motion blur on DTB70.]
	{
		\includegraphics[width=0.24\linewidth]{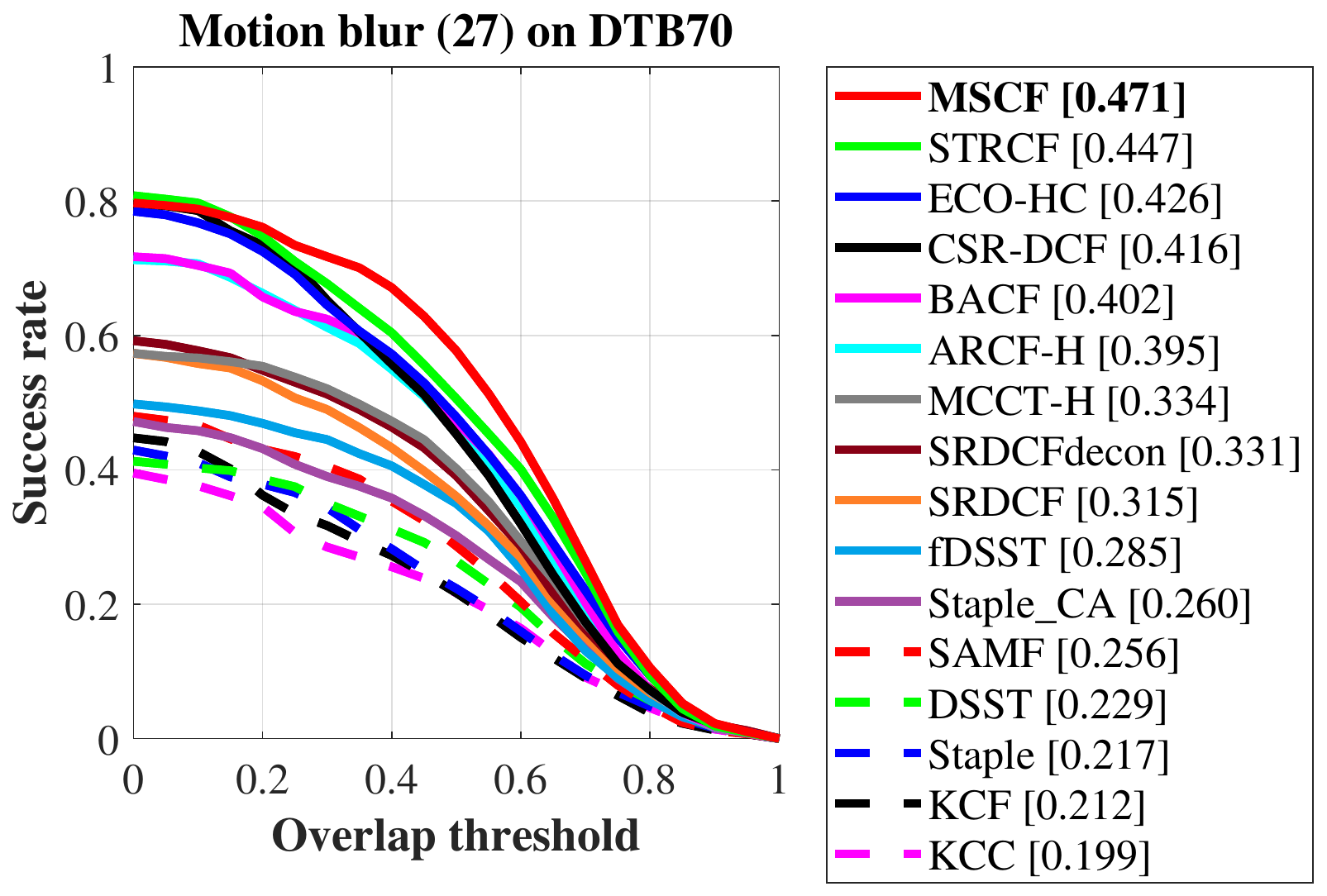}
	}
	\caption
	{
		Attribute-based performance evaluation. Success plots of MSCF and other state-of-the-art trackers on UAV-specific attributes from three benchmarks. MSCF outperforms other trackers on all challenges above.
	}
	\label{fig:att_hand}
\end{figure*}

\subsection{Comparison with CPU-based Trackers}
\subsubsection{Overall performance} Fig.~\ref{fig:hand_over} demonstrates the overall performance of MSCF with other trackers on UAV123@10fps~\cite{Mueller2016ECCV}, DTB70~\cite{Li2017AAAI}, and UAVDT~\cite{Du2018ECCV} benchmarks. The proposed MSCF tracker has outperformed all other trackers with handcrafted features on three benchmarks. On UAV123@10fps, MSCF ranks first in precision (0.648) and AUC (0.471). While on UAVDT, MSCF (0.713) has an advantage of 1.1\% over the second-best ARCF-H (0.705) on precision. In terms of AUC, MSCF (0.455) has an advantage of 5.4\% over the second-best BACF (0.432). On DTB70 benchmark, MSCF has also achieved the best performance in precision (0.664) and AUC (0.459). Overall evaluation of performance on all three benchmarks in terms of precision, AUC, and FPS is demonstrated in Table~\ref{tab:hand_avg}. Against baseline BACF, MSCF has an advancement of 10.1\% in precision and 11.6\% in AUC. Besides satisfactory tracking results, the speed of MSCF (37.6 FPS) is adequate for real-time UAV tracking applications.
\subsubsection{Attribute-based performance}
\begin{figure}[!t]	
	\centering
	\includegraphics[width=0.46\textwidth]{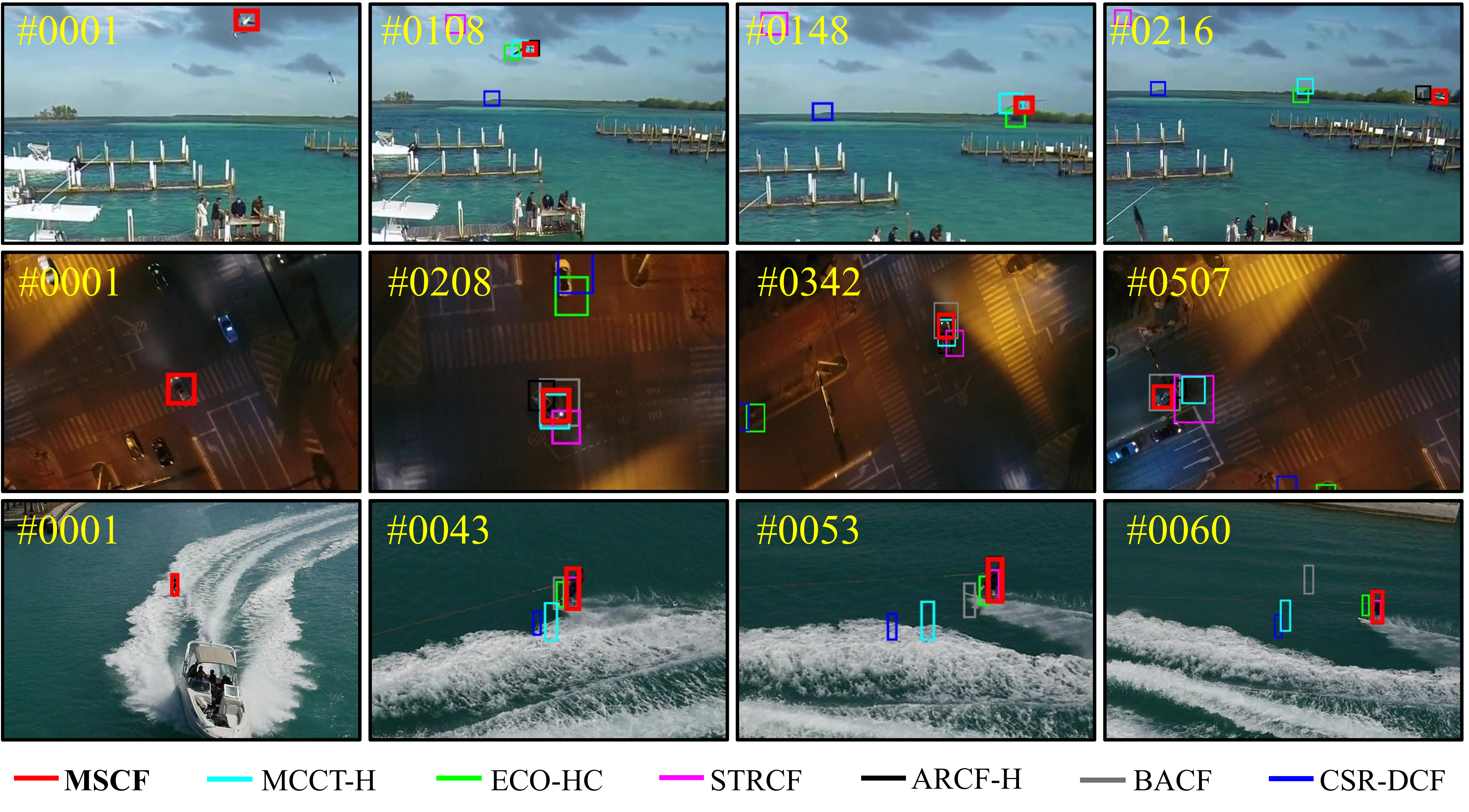}
	\setlength{\abovecaptionskip}{-0.2cm}
	\caption
	{Qualitative evaluation of 7 top-ranked trackers. From top to down are $Gull2$ from DTB70, $S1301$ from UAVDT, and $wakeboard7$ from UAV123@10fps. More tracking videos can be found here: \url{https://youtu.be/4EPGBCuatxU}.
	}
	\label{fig:quan}
\vspace{-0.2cm}
\end{figure}

SPs of all trackers on UAV-specific attributes of three benchmarks are demonstrated in Fig.~\ref{fig:att_hand}. The figure illustrates that mutations caused by challenges including fast motion, camera motion, \textit{etc.}, are commendably sensed and repressed by the MSCF tracker. In case of camera motion, MSCF demonstrates an improvement of performance by 14.6\% in UAVDT, and 13.1\% in DTB70 against BACF. The ability to cope with object motion has also been promoted by 20.7\% in UAV123@10fps, 7.3\% in UAVDT and 17.2\% in DTB70 from BACF. For explication, mutations of object or camera motion commonly bring sub-peaks in response maps, and the proposed MTF can instantly correct the hybrid label for filter training. Taking the neighborhood of the centre into consideration, the label can adapt to the mutation of the appearance smoothly. Some qualitative comparisons of the top 8 trackers are displayed in Fig.~\ref{fig:quan}.

\begin{table}[!t]
	\centering
	\caption{Performance comparison of MSCF with other 11 state-of-the-art deep-based trackers on UAVDT benchmark. Note that the superscript * denotes GPU speed.}
	\vspace{-9pt}
	\resizebox{\linewidth}{!}{
	\begin{tabular}{lccc||lccc}
		\hline
		\hline
		Trackers & Prec.  & Succ.  & FPS    & Trackers & Prec.  & Succ.  & FPS \\
		\hline
		CCOT   & 0.656  & 0.406  & 1.1*   & DeepSTRCF & 0.667  & 0.437  & 6.8* \\
		CoKCF  & 0.605  & 0.319  & 20.1*  & TADT   & 0.677  & 0.431  & 32.3* \\
		ECO    & \textcolor[rgb]{ 0,  0,  1}{\textbf{0.700}} & \textcolor[rgb]{ 0,  1,  0}{\textbf{0.454}} & 16.4*  & UDT    & 0.674  & \textcolor[rgb]{ 0,  0,  1}{\textbf{0.442}} & \textcolor[rgb]{ 1,  0,  0}{\textbf{73.3}}* \\
		IBCCF  & 0.603  & 0.389  & 3.0*   & UDT+   & 0.696  & 0.415  & \textcolor[rgb]{ 0,  1,  0}{\textbf{56.9}}* \\
		MCPF   & 0.675  & 0.403  & 0.6*   & ASRCF  & \textcolor[rgb]{ 0,  1,  0}{\textbf{0.700}} & 0.437  & 22.2* \\
		MCCT   & 0.671  & 0.437  & 7.9*   & \bf{MSCF} & \textcolor[rgb]{ 1,  0,  0}{\textbf{0.713}} & \textcolor[rgb]{ 1,  0,  0}{\textbf{0.455}} & \textcolor[rgb]{ 0,  0,  1}{\textbf{37.6}} \\
		\hline
		\hline
	\end{tabular}}
	\label{tab:deep}%
	\vspace{-9pt}
\end{table}%

\begin{table}[!t]
	\centering
	\caption{Ablation study on average performance of the proposed MSCF tracker.}
	\vspace{-9pt}
\begin{tabular}{lcccc}
\hline
\hline
Settings & BACF & BACF+AL & BACF+AHL & \textbf{MSCF} \\
\midrule
Prec. & 0.613 & \textcolor[rgb]{ 0,  0,  1}{\textbf{0.668}} & \textcolor[rgb]{ 0,  1,  0}{\textbf{0.673}} &\textcolor[rgb]{ 1,  0,  0}{\textbf{0.675}} \\
Succ. & 0.414 & \textcolor[rgb]{ 0,  0,  1}{\textbf{0.456}} & \textcolor[rgb]{ 0,  1,  0}{\textbf{0.458}} &\textcolor[rgb]{ 1,  0,  0}{\textbf{0.462}} \\
\hline
\hline
\end{tabular}%
\label{tab:ablation}%
\end{table}%
\subsection{Comparison with Deep-Based Trackers }
To better illustrate the comprehensive evaluation with state-of-the-art trackers, the performance of 11 deep-based methods are also represented in comparison with MSCF in TABLE~\ref{tab:deep}. Owing to heavy reliance on high-end GPUs for deep-based computation, while the MSCF tracker is merely in need of a single CPU, the prime cost of MSCF is accordingly lower than deep-based trackers. Moreover, the proposed MSCF tracker is superior to deep-based trackers on both precision and success rate. 
\subsection{Albation Study}
TABLE~\ref{tab:ablation} demonstrates the availability of every amendment for the baseline tracker. As the adaptive label (AL), \textit{i.e.}, the first two terms in Eq.~(\ref{equ: main}), is applied in the objective function, where the predefined Gaussian label is replaced by a trained label, the precision and success rate have achieved an improvement by 9.0\% and 10.1\%, respectively. Taking the adaptive hybrid label (AHL) into consideration, \textit{i.e.}, appending the spatial regularization term to the objective function, the performance has a further improvement. Moreover, after utilizing the AHL in the previous frame, \textit{i.e.}, appending the temporal regularization term, the complete MSCF has been promoted by 0.3\% in precision and 0.9\% in success rate. To sum up, the MSCF has gained superior competence on the baseline.
\section{Conclusion}

In this work, a mutation sensitive tracker with an adaptive hybrid label is proposed for real-time UAV tracking. Meanwhile, a novel evaluation of the response map is introduced, namely MTF, which indicates the severity of tracking mutations. The whole work-flow also adopts the genetic idea to reach optimal solutions. Comprehensive experiments have corroborated the favorable performance of MSCF. Furthermore, employing the visual tracking adaptability of MSCF for UAV tracking in the real environment is also taken into consideration to conduct in the future. We believe that DCF based framework for real-time UAV tracking can be further improved with the proposition of our MSCF tracker.






\section*{Acknowledgement}
This work is supported by the National Natural Science Foundation of China (No. 61806148) and the Natural Science Foundation of Shanghai under Grant (No. 20ZR1460100).

\bibliographystyle{IEEEtran}
\normalem
\bibliography{root.bib}

\end{document}